\PassOptionsToPackage{hyperfootnotes=false}{hyperref}
\documentclass[11pt,a4paper,logo,copyright]{xiaomi}

\usepackage[numbers,sort&compress,square]{natbib}
\usepackage{amsmath,amssymb,amsfonts}
\usepackage{booktabs,array,tabularx}
\usepackage{graphicx}
\usepackage{colortbl}
\usepackage{placeins}
\usepackage{float}


\definecolor{claimblue}{RGB}{235,245,255}
\definecolor{teaserSC}{RGB}{33,95,154}
\definecolor{teaserDD}{RGB}{192,0,0}
\definecolor{teaserNT}{RGB}{19,80,27}
\definecolor{LakeBlue}{RGB}{0,61,153}
\hypersetup{
  colorlinks=true,
  allcolors=LakeBlue,
  pdftitle={Tether the Subject, Release the Scene: Query-Aware Memory Routing for Long-Horizon Autoregressive Video Generation},
  pdfauthor={Chen Li, Peng Zhang, Hanyu Zhou, Jialong Zuo, Fei Wang, Daiguo Zhou, Nong Sang, Changxin Gao}
}
\newcommand{\decnum}[2]{\makebox[0.75em][r]{#1}\kern0.04em.\kern0.04em\makebox[1.15em][l]{#2}}

\title{\centering Tether the Subject, Release the Scene: Query-Aware Memory Routing for Long-Horizon Autoregressive Video Generation}
\titlerunning{}
\authorrunning{Chen Li et al.}
\author{Chen Li\textsuperscript{1,2}, Peng Zhang\textsuperscript{2},
Hanyu Zhou\textsuperscript{1}, Jialong Zuo\textsuperscript{1},\\
Fei Wang\textsuperscript{2}, Daiguo Zhou\textsuperscript{2},
Nong Sang\textsuperscript{1}, Changxin Gao\textsuperscript{1}\thanks{Corresponding author.}}
\institute{\textsuperscript{1}Huazhong University of Science and Technology\\
\textsuperscript{2}MiLM Plus, Xiaomi Inc.\\[4pt]
Project page: \href{https://lichen1015.github.io/tethermem/}{https://lichen1015.github.io/tethermem/}}

\fancypagestyle{firststyle}{
  \fancyhead[L,R,C]{}
  \fancyhead[L]{\raisebox{-0.5pt}{\includegraphics[height=23pt]{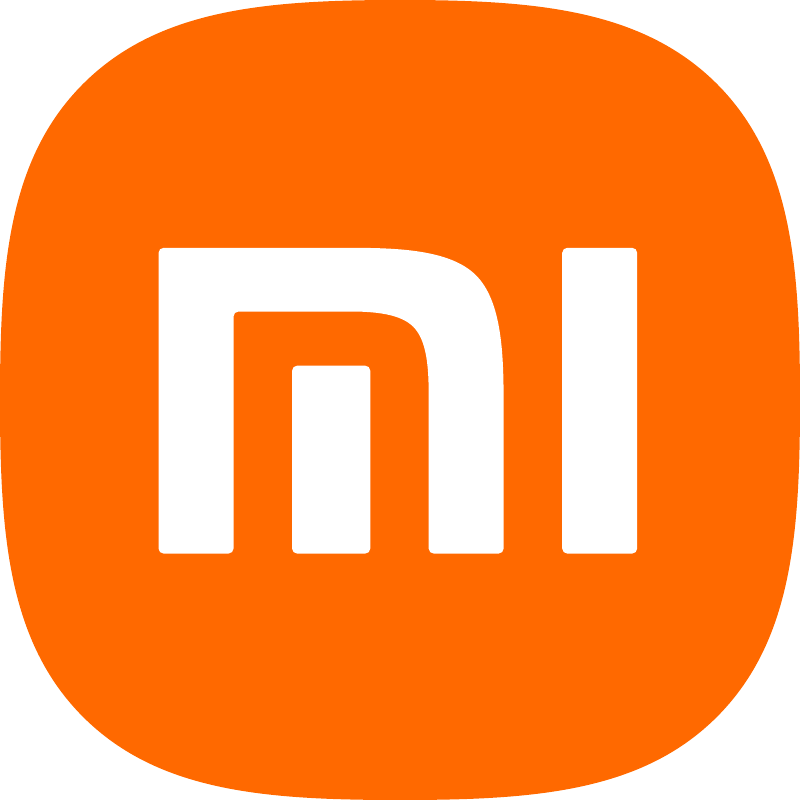}}}
  \fancyhead[R]{\raisebox{-0.5pt}{\includegraphics[height=23pt]{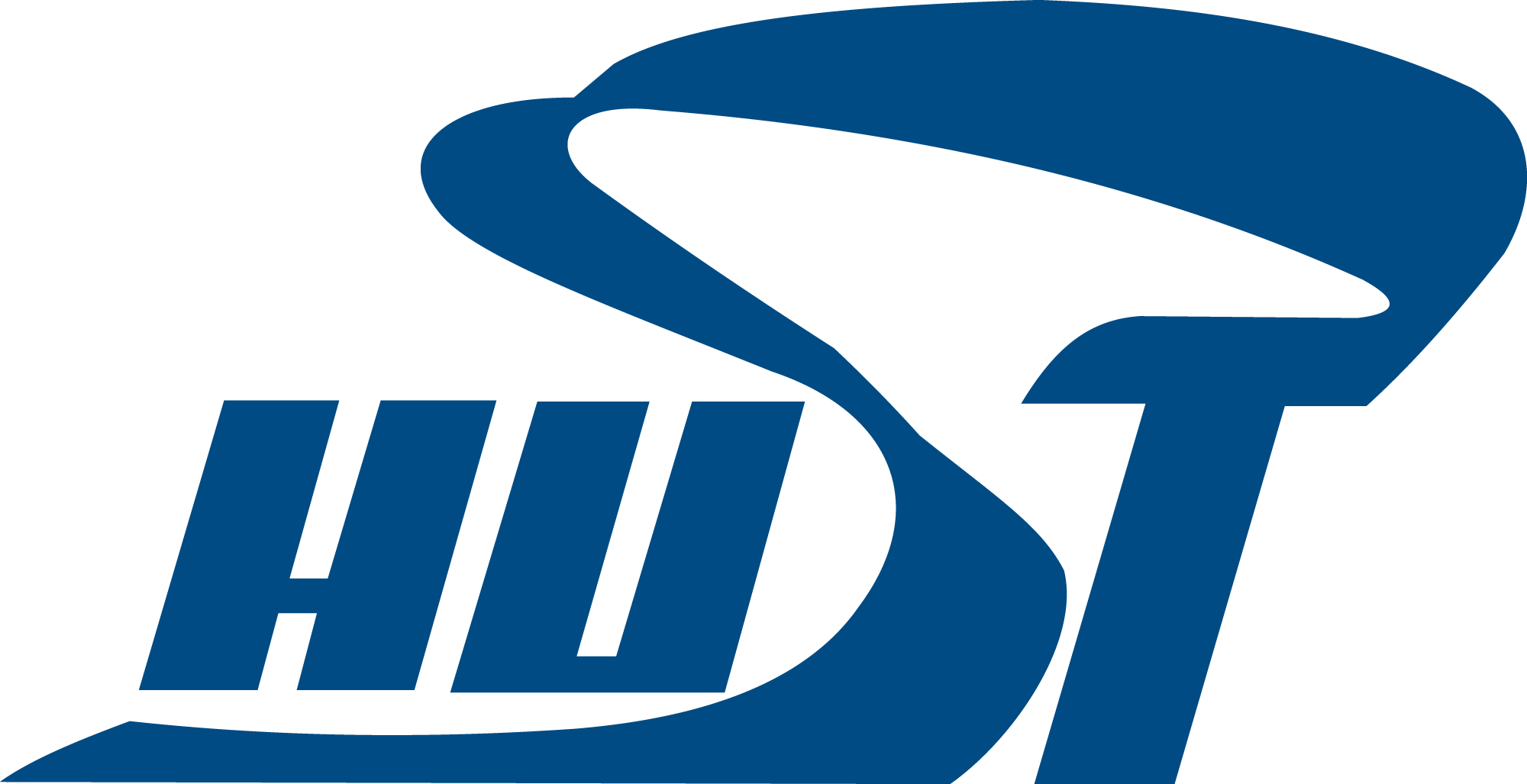}}}
  \fancyfoot[C]{}
}

\begin{document}

\begin{abstract}
Streaming autoregressive video models generate long videos chunk by chunk, using
historical memory to maintain consistency. Existing methods typically expose
subject and scene queries to history through similar policies. This stabilizes
the subject, but can also lock backgrounds, viewpoints, and scene structure to
previously generated states even when local motion continues. We call this
failure \emph{memory-anchored scene under-progression}; consistency and motion
metrics alone can miss it. We introduce TetherMem, a training-free, query-aware
spatiotemporal memory router for frozen video generators. TetherMem separates
subject and scene queries and modulates historical access with region- and
age-conditioned priors: subject queries retain identity-bearing history, while
scene queries reduce reliance on subject history and stale backgrounds. Across
2,400 blinded pairwise judgments from 10 annotators, TetherMem achieves the
highest estimated expected preference among eight streaming long-video
baselines for overall quality (0.780) and scene progression (0.769). On complete
30-second videos, it sustains changes in background, viewpoint, and scene state
while preserving subject recognizability and temporal continuity.
\end{abstract}

\maketitle

\begin{figure}[t]
  \centering
  \includegraphics[width=\textwidth]{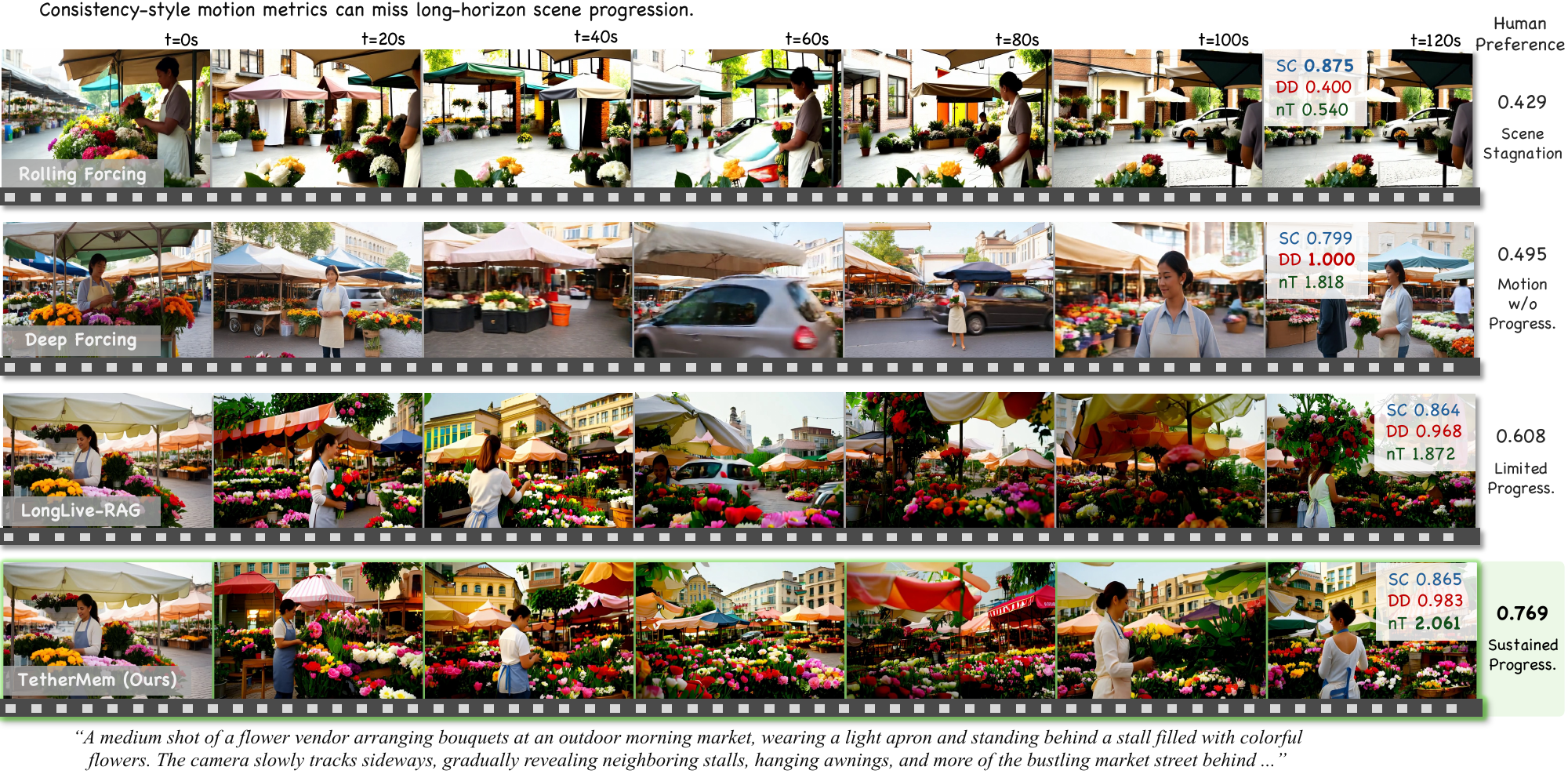}
  \caption{\textbf{Memory-anchored scene under-progression.}
  This matched prompt--seed trajectory illustrates distinct long-horizon scene
  dynamics across four generation methods, with TetherMem continuing to reveal
  new scene states through 120 seconds. Human Preference reports the aggregate
  30-second scene-progression expected preference from
  Table~\ref{tab:main_sota}. VBench Subject Consistency
  (\textcolor{teaserSC}{SC}) measures subject preservation, Dynamic Degree
  (\textcolor{teaserDD}{DD}) captures overall motion, and
  \textcolor{teaserNT}{nT} is an auxiliary scene-progression diagnostic.}
  \label{fig:tethermem_overview}
\end{figure}

\section{Introduction}
\label{sec:introduction}

Streaming autoregressive (AR) video diffusion models have become an important
paradigm for long-horizon video generation
\cite{yin2024causvid,huang2025selfforcing,yang2025longlive,liu2025rollingforcing}.
They decompose long-video synthesis into a causal sequence of chunks, each
conditioned on previously generated content. This chunk-wise process enables
real-time and continuously extendable generation, but it also allows errors to
accumulate over time. Subject identity, geometry, and scene layout may drift as
generation proceeds. Existing methods therefore rely heavily on historical
memory, carrying past content into future chunks through persistent key--value
caches or positional memory~\cite{liu2025rollingforcing,kim2026memrope},
attention anchors~\cite{yang2025longlive}, and retrieval-based
memory~\cite{hu2026longliverag,yu2025contextmemory}.
Historical memory has consequently become a central component for maintaining
long-video consistency.

However, while suppressing forgetting, historical memory can also impede the
transition to new scene states. When subject and scene queries repeatedly
access previously generated content through similar policies, memory not only
protects subject identity but also anchors the background, viewpoint, and scene
structure to the past. As shown in Figure~\ref{fig:tethermem_overview},
long-memory methods can generate videos in which
the subject remains stable and local motion persists, yet the background and
viewpoint remain confined to repetitions or slight extensions of states that
have already appeared. The scene progression requested by the prompt fails to
materialize. We call this failure \textbf{memory-anchored scene
under-progression}. Unlike overt identity drift, it can coexist with a stable
subject and local motion, motivating evaluation of whether the scene continues
to progress over the full video.

Other approaches weaken, update, or evolve historical states to restore
dynamics~\cite{dalva2026adastate,bian2026echoinfinity,zhao2026relaxforcing}.
These methods typically relax memory at the frame or global level: dynamics
improve, but subject identity and structure can weaken. They still apply
approximately the same access policy to subject and scene content, despite
their different roles. The subject should preserve its identity, whereas the
background, viewpoint, and spatial relations should continue toward new states.
A global memory policy therefore trades stability for progression.

A natural way to break this trade-off is to attenuate historical Values in
background regions after attention. Although this releases more scene
variation, it leaves the original attention routing unchanged and introduces a
query-dependent scale on the attention output, which can weaken subject
identity and structure. Moving the spatial prior into the attention logits
instead redistributes historical access within the normalized attention
distribution. Spatial routing alone, however, still assigns every current
query the same policy. This leads to the central question: \emph{which history
should each query retrieve to preserve the subject while allowing the scene to
progress?}

To address this question, we propose \textbf{TetherMem}, a training-free,
query-aware spatiotemporal memory routing method that can be directly applied
to a frozen video generator, given frame-wise subject priors. TetherMem
distinguishes subject queries from scene queries and controls their access to
history through region- and memory-age-conditioned routing priors. Subject
queries preferentially retrieve identity-related history, whereas scene queries
reduce their dependence on historical subject content and stale background
states. These priors act on the attention logits without modifying historical
Values, allowing each query type to retrieve evidence suited to its needs and
thereby balance subject stability with scene progression. Across eight streaming
long-video baselines, TetherMem attains the highest estimated expected
preference (EP) for overall quality and scene progression (0.780 and 0.769).

Our contributions are summarized as follows.
\begin{itemize}
  \item We identify \textbf{memory-anchored scene under-progression} and
  distinguish sustained scene progress from subject stability and local motion.
  \item We introduce training-free query-, region-, and age-conditioned memory
  routing, with normalized selection in place of post-attention Value Reweighting.
  \item Across eight streaming long-video baselines, TetherMem leads estimated
  overall-quality and scene-progression EP in blinded human comparisons.
\end{itemize}

\section{Related Work}
\label{sec:related_work}

\textbf{Historical memory in long-video generation.}
Long-horizon autoregressive video generation commonly relies on historical
states to limit error accumulation. Existing methods strengthen this capability
along three main directions. The first builds more persistent memory carriers.
LongLive uses frame-level attention sinks to preserve long-range
context~\cite{yang2025longlive}. Rolling Forcing retains key--value states from
initial frames as global context anchors~\cite{liu2025rollingforcing}.
Infinity-RoPE extends long-context generation through positional encoding and
KV-cache mechanisms~\cite{yesiltepe2025infinityrope}.
MemRoPE and retrieval-based methods instead manage historical content through
compression, updating, or retrieval on demand
\cite{kim2026memrope,hu2026longliverag,yu2025contextmemory}. The second
direction improves the training objective. CausVid and Self Forcing use
distillation, self-generated rollouts, or distribution matching to reduce
train--test mismatch and long-horizon error accumulation
\cite{yin2024causvid,huang2025selfforcing}. A third direction relaxes or evolves
historical states to recover long-range dynamics, as in AdaState and Deep
Forcing~\cite{dalva2026adastate,yi2025deepforcing}. These approaches improve
memory duration, capacity, update rules, or training. TetherMem complements
them by routing the available historical memory according to query role, key
region, and memory age, creating distinct access paths for identity preservation
and scene progression.

\textbf{Selective caching and memory routing.}
In language models, H$_2$O and SnapKV retain selected KV-cache tokens using
attention-derived importance or query-dependent prompt features
\cite{zhang2023h2o,li2024snapkv}. Their central operation is cache selection for
computational efficiency. Diffusion editing methods also manipulate cross- or
self-attention to preserve spatial layout and appearance
\cite{hertz2023prompt,cao2023masactrl}. TetherMem instead addresses
query-specific access to history
during long-video generation. Subject queries require stable access to
identity-bearing evidence, whereas scene queries should reduce their dependence
on subject history and stale background states. Region and age priors in the
attention logits organize this access according to each query's generation
role while leaving historical Values unchanged.

\section{Problem Formulation}
\label{sec:problem}
Long videos are generated autoregressively, one chunk at a time. To generate
the $n$-th video chunk, the model is conditioned on a text prompt $P$, random
noise $\epsilon_n$, and historical memory $M_n$:

\begin{align}
X_n &= G_\theta(\epsilon_n; P, M_n).
\end{align}

Here, $M_n$ contains historical key--value states obtained by caching or
retrieval. Let $q\in\mathcal{Q}_n$ be a query token in the current chunk, and
let $\mathcal{C}_n$ be the set of context tokens accessible to it. The subset
$\mathcal{I}_n\subset\mathcal{C}_n$ contains the historical tokens; the
remaining tokens provide local context or act as persistent attention sinks.
The base attention is

\begin{align}
A_{qi}
&=
\frac{
\exp\!\left(Q_qK_i^\top/\sqrt{d}\right)
}{
\sum_{j\in\mathcal{C}_n}
\exp\!\left(Q_qK_j^\top/\sqrt{d}\right)
},
\label{eq:base_attention} \\
O_q
&=
\sum_{i\in\mathcal{C}_n} A_{qi}V_i.
\label{eq:base_output}
\end{align}

where $Q_q$ is the current query representation, $(K_i,V_i)$ is the key--value
pair for token $i$, and $d$ is the feature dimension. Historical tokens provide
cross-chunk continuity for subject identity, local structure, and scene content.
Repeatedly accessing the same historical states, however, can also anchor the
background, viewpoint, and scene layout to content that has already appeared.

To explicitly control access to history, we introduce a positive
routing prior $\pi_n(q,i)>0$:

\begin{align}
\widetilde{A}_{qi}
&=
\frac{
\pi_n(q,i)
\exp\!\left(Q_qK_i^\top/\sqrt{d}\right)
}{
\sum_{j\in\mathcal{C}_n}
\pi_n(q,j)
\exp\!\left(Q_qK_j^\top/\sqrt{d}\right)
},
\label{eq:controlled_attention}
\end{align}

\begin{figure*}[t]
\centering
\includegraphics[width=\textwidth]{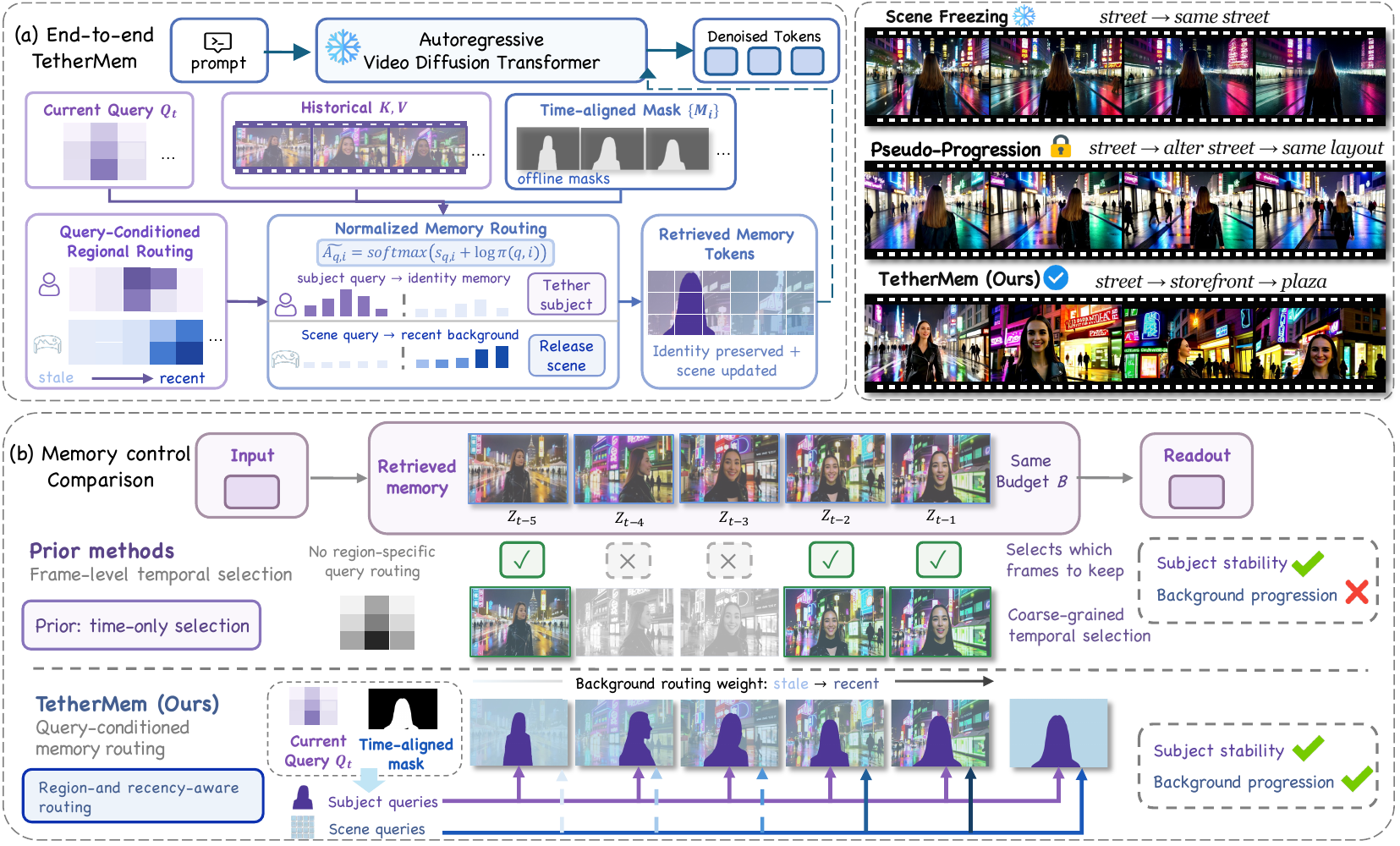}
\caption{\textbf{Overview of TetherMem.}
\textbf{(a)} End-to-end routing preserves identity memory for subject queries
and favors recent backgrounds for scene queries. \textbf{(b)} Prior methods
select memory frames, whereas TetherMem routes query roles within the retrieved history.}
\label{fig:tethermem_architecture}
\end{figure*}

This prior adjusts the relative weight with which the current query accesses
different historical tokens. For non-historical tokens, we set
$\pi_n(q,i)=1$ so that local context and attention-sink tokens retain their
original behavior. When the prior equals one everywhere, Eq.~\eqref{eq:controlled_attention}
reduces to the base attention. Existing controls mainly operate on cache
length, retrieval scope, frame weights, or overall memory strength. TetherMem
exposes query-role conditioning at the same attention interface, allowing
subject and scene queries to follow different policies for accessing history.

The problem with this shared policy is that subject and scene generation require
different historical information. Subject generation must repeatedly access
identity- and structure-bearing cues. Scene generation, in contrast, must
reduce its dependence on stale backgrounds so that viewpoints, spatial
relations, and environmental content can continue to evolve. Applying the same
historical constraint to both therefore creates a trade-off between stability
and progression: \textbf{stronger historical access can cause the scene to
stagnate, whereas weaker historical access can undermine subject stability.}

This paper therefore asks: \emph{Should subject and scene queries access
historical memory differently to jointly support identity preservation and
scene progression?}

\section{TetherMem}
\label{sec:method}

Figure~\ref{fig:tethermem_architecture} overviews TetherMem, which conditions
historical access on the current-query role, historical-key region, and memory
age. The binary subject mask operationalizes scene queries as non-subject, or
background, tokens. Normalized routing places the prior inside the softmax so
that it changes which history is selected without introducing an additional
output scale. Regional routing then conditions access on the query--key region,
while recency routing directs background queries toward recent scene states and
preserves the long-range subject path. Secs.~\ref{sec:normalized_routing}--
\ref{sec:recency_routing} develop these three components in order.

\subsection{Normalized Memory Routing}
\label{sec:normalized_routing}

Subject identity requires historical evidence, whereas scene progression
requires weaker dependence on previously observed backgrounds. A direct
approach is to preserve historical Values in subject regions while attenuating
those in background regions. Following the notation of
Sec.~\ref{sec:problem}, let $w_i\in(0,1]$ be a spatial retention weight, with
larger values for historical subject regions and smaller values for historical
background regions. We call this post-attention spatial modulation
\emph{Value Reweighting}:

\begin{align}
O_q^{\mathrm{val}}
&=
\sum_{i\in\mathcal{C}_n} A_{qi}w_iV_i.
\label{eq:value_reweighting}
\end{align}

For local and sink tokens outside the retrieved history, we set $w_i=1$. This
modulation can release some background variation, but it is applied only after
attention has selected the information. Its effect can be decomposed as

\begin{align}
O_q^{\mathrm{val}}
&=g_q\sum_{i\in\mathcal{C}_n}\overline{A}_{qi}V_i,
\label{eq:value_decomposition}\\
g_q
&=\sum_{i\in\mathcal{C}_n}A_{qi}w_i,
\label{eq:value_gain}\\
\overline{A}_{qi}
&=\frac{A_{qi}w_i}{g_q},
\qquad
\sum_{i\in\mathcal{C}_n}\overline{A}_{qi}=1.
\label{eq:renormalized_attention}
\end{align}

Here, $\overline{A}_{qi}$ is the renormalized contribution after spatial
reweighting and satisfies $\sum_i\overline{A}_{qi}=1$, while $g_q$ is an
additional output scale factor. Because queries have different base attention
distributions, the shared spatial weights $w_i$ produce different values of
$g_q$. Value Reweighting therefore not only reallocates the relative
contributions of historical evidence, but also introduces a query-dependent
perturbation to the magnitude of the attention output, which is repeatedly fed
into subsequent history and can accumulate during a long rollout.

This observation motivates the normalized routing used by TetherMem:
\textbf{the spatial prior should participate in selecting historical
information, rather than scaling retrieved Values after selection.} We
therefore move spatial control inside the attention softmax. Let
$s_{qi}=Q_qK_i^\top/\sqrt d$ denote the original attention logit, and introduce
a positive routing prior $\pi_n(q,i)>0$. The controlled attention is

\begin{align}
\widetilde{A}_{qi}
&=
\frac{\pi_n(q,i)\exp(s_{qi})}
{\sum_{j\in\mathcal{C}_n}\pi_n(q,j)\exp(s_{qj})},
\label{eq:normalized_routing}\\
\widetilde{O}_q
&=
\sum_{i\in\mathcal{C}_n}\widetilde{A}_{qi}V_i.
\label{eq:normalized_output}
\end{align}

Equivalently, $\log\pi_n(q,i)$ acts as an additive bias on the original
attention logit during softmax normalization. Thus,
$\widetilde{A}_{qi}$ always satisfies
$\sum_i\widetilde{A}_{qi}=1$, while the historical Values remain unchanged.
TetherMem therefore changes which historical locations each query reads from,
rather than scaling retrieved content afterward.

\subsection{Query-Conditioned Regional Routing}
\label{sec:regional_routing}

The role of a historical token is determined jointly by the query and key
regions. A historical subject token provides identity and structural evidence
for a subject query, while the same token can anchor a background query to an
existing layout. Historical backgrounds maintain scene continuity when read by
background queries, while cross-region access receives a softer prior.

TetherMem therefore routes memory according to the regional relation between
the current query and each historical key. Let
$m^q(q)$ and $m^k(i)$ indicate whether the current query $q$ and historical key
$i$, respectively, lie in the subject regions of their corresponding frames:
\begin{align}
m^q(q),m^k(i)&\in\{0,1\},
\end{align}
where one denotes the subject region and zero denotes the background region.
We instantiate this binary prior by generating a Full-Memory reference video,
extracting a frame-wise subject track with SAM~2~\cite{ravi2024sam2}, and
downsampling the track to the latent-token grid. Each query and historical key
uses the reference mask at its current or source time, respectively; this
time-aligned track is the routing scaffold. Appendix~\ref{sec:prior_alignment_audit}
measures its alignment over the rollout.

For a key $i\in\mathcal{I}_n$ in the retrieved history, TetherMem assigns a unit
prior when the query and key belong to the same region, leaving the connection
unattenuated before attention normalization. When they belong to different
regions, it downweights the connection using a cross-region routing factor
$\gamma_n\in(0,1]$:

\begin{align}
\pi_{\mathrm{reg}}(q,i)
&=
\begin{cases}
1, & m^q(q)=m^k(i),\\
\gamma_n, & m^q(q)\neq m^k(i),
\end{cases}
\qquad 0<\gamma_n\leq1.
\label{eq:regional_prior}
\end{align}

Here, $\gamma_n=1$ applies no regional control, while a smaller $\gamma_n$
suppresses cross-region access. We recompute $\gamma_n$ from the current
subject fraction $r_n$ using the fixed release budget $\alpha=0.25$:
\begin{align}
\bar\gamma_n&=\frac{\alpha-r_n}{1-r_n},
&
\gamma_n&=\max\!\left(10^{-9},\min(1,\max(0,\bar\gamma_n))\right).
\label{eq:gamma_explicit}
\end{align}
This area-calibrated rule keeps the cross-region release budget consistent as
subject area changes; we use the fixed $\alpha=0.25$ throughout. The all-subject
edge case is given in Appendix~\ref{sec:implementation_details}.
Substituting
$\pi_{\mathrm{reg}}(q,i)$ into the normalized attention in
Sec.~\ref{sec:normalized_routing} yields query-conditioned regional memory
routing. For local and sink tokens, the routing prior remains one.

This design creates two complementary paths for accessing history. Subject
queries retain strong access to historical subject evidence while reducing
interference from background history. Background queries preferentially read
historical backgrounds and reduce their dependence on subject history,
preventing the scene from being continually anchored to existing subject
layouts. The key is to make the importance of historical evidence depend on the
generation role of the current query, rather than assigning a fixed importance
to each historical token in advance.

\subsection{Recency-Aware Background Routing}
\label{sec:recency_routing}

Regional routing selects the memory type; recency routing further orders states
within each type. Earlier viewpoints, spatial layouts, or completed scene
states may regain high attention when they resemble the current content,
causing the background to stall, return, or slowly repeat.

Recent background memory more closely reflects the current scene state and
helps maintain continuity between adjacent chunks. Older background memory is
more likely to describe a state that the video has already left. Based on this
observation, TetherMem introduces a recency prior for historical backgrounds.
Let $p_i$ be the source-frame position of historical token $i$ in the memory
pool and $A_{\max}=\max(1,\min(N_{\mathrm{pool}},120))$. Its normalized
absolute recency is $\tau_i=p_i/A_{\max}$, with larger values indicating states
closer to the current time. We define

\begin{align}
\rho_i&=\max(\tau_i,\rho_{\min}),
\label{eq:recency_prior}
\end{align}

where the fixed floor $\rho_{\min}=0.05$ retains a small contribution from
distant background states.

Combining this prior with the regional routing in
Sec.~\ref{sec:regional_routing}, the complete prior for a historical key
$i\in\mathcal{I}_n$ is

\begin{align}
\pi_n(q,i)
&=
\begin{cases}
1, & m^q(q)=1,\;m^k(i)=1,\\
\gamma_n, & m^q(q)\neq m^k(i),\\
\rho_i, & m^q(q)=0,\;m^k(i)=0.
\end{cases}
\label{eq:complete_prior}
\end{align}

For local and sink tokens, we continue to set $\pi_n(q,i)=1$. Subject queries
therefore retain long-range identity cues, while background queries favor recent
scene states.

\section{Experiments}
\label{sec:experiments}

\subsection{Experimental Setup}
\label{sec:experimental_setup}

We evaluate ten prompts spanning stable scenes (P01--P03),
subject-centered progression (P04 and P06--P10), and subject-free evolution
(P05), each rendered with three seeds. Appendix~\ref{sec:prompts_protocol}
gives the complete prompts and executed settings.

Wan2.1-T2V-1.3B~\cite{wanteam2025wan} serves as a 5-second base-model
reference. TetherMem is built on LongLive-RAG~\cite{hu2026longliverag} and uses
one routing configuration. Using officially released weights, we
compare against eight streaming long-video baselines: LongLive-RAG, Self
Forcing, Rolling Forcing, Deep Forcing, Causal Forcing, Reward Forcing,
MemRoPE, and CausVid
\cite{hu2026longliverag,huang2025selfforcing,liu2025rollingforcing,
yi2025deepforcing,zhu2026causalforcing,lu2025rewardforcing,
kim2026memrope,yin2024causvid}. The controlled evaluation uses this
Wan2.1-T2V-1.3B/LongLive-RAG stack, ten prompts, three seeds, and approximately
30-second videos. Longer 42-, 52-, and 120-second rollouts illustrate the same
long-horizon behavior.

Human evaluation contains 2,400 blinded pairwise judgments from 10 independent
annotators on complete 30-second videos. Randomized pairs are rated for overall
preference (Ovr.), scene progression, subject identity preservation (ID), and
visual integrity (L1). Tie-aware Davidson--Bradley--Terry models place methods on
a common EP scale~\cite{davidson1970ties}; the rubric, model, uncertainty, and
quality-control details are in Appendix~\ref{sec:human_stats}.

Overall preference and scene progression on the complete 30-second videos are
the primary criteria. Identity is evaluated on the nine subject-present
prompts; P05 is marked N/A. Img5 is VBench Imaging
Quality~\cite{huang2023vbench} on 5-second clips and checks short-horizon
degradation. nT measures net coherent translation from optical flow and serves
as an auxiliary cue for sustained directional change.

\subsection{Main Comparison}
\label{sec:main_quant_results}

\begin{table}[!t]
\centering
\begingroup
\small
\setlength{\tabcolsep}{3.0pt}
\renewcommand{\arraystretch}{1.03}
\begin{tabularx}{\textwidth}{l*{8}{>{\centering\arraybackslash}X}c>{\centering\arraybackslash}X}
\toprule
Method
& \multicolumn{1}{c}{\textbf{5s}}
& \multicolumn{4}{c}{\textbf{30s human EP}$\uparrow$}
& \multicolumn{1}{c}{\textbf{Diag.}}
& \multicolumn{4}{c}{\textbf{VBench}$\uparrow$} \\
\cmidrule(lr){2-2}\cmidrule(lr){3-6}\cmidrule(lr){7-7}\cmidrule(lr){8-11}
& Img5$\uparrow$ & Ovr. & Prog. & ID & L1 & nT$\uparrow$
& Subj. & Back. & Smooth. & Dyn. \\
\midrule
\rowcolor{black!5}\multicolumn{11}{l}{\textit{\textbf{Base model (non-autoregressive)}}}\\
Wan2.1-T2V-1.3B
& 65.6 & -- & -- & -- & -- & 0.322 & -- & -- & -- & -- \\

\rowcolor{black!5}\multicolumn{11}{l}{\textit{\textbf{Long-memory / retrieval}}}\\
LongLive-RAG
& 71.0 & 0.615 & \underline{0.608} & 0.507 & 0.587 & 0.868
& 0.861 & 0.897 & 0.973 & \textbf{0.867} \\

\rowcolor{black!5}\multicolumn{11}{l}{\textit{\textbf{Forcing-based generation}}}\\
Self Forcing
& 70.5 & 0.320 & 0.356 & 0.437 & 0.273 & 0.192
& 0.885 & 0.907 & 0.983 & 0.767 \\
Rolling Forcing
& \textbf{72.2} & 0.553 & 0.429 & \textbf{0.633} & 0.651 & 0.040
& \textbf{0.933} & \textbf{0.935} & \underline{0.985} & 0.433 \\
Deep Forcing
& 70.5 & 0.526 & 0.495 & 0.531 & 0.508 & 0.350
& 0.899 & 0.916 & 0.983 & 0.800 \\
Causal Forcing
& 70.2 & 0.210 & 0.440 & 0.285 & 0.158 & \underline{1.226}
& 0.824 & 0.873 & 0.973 & \textbf{0.867} \\
Reward Forcing
& 69.3 & 0.573 & 0.530 & 0.487 & 0.620 & 0.366
& \underline{0.903} & \underline{0.927} & \underline{0.985} & 0.733 \\
CausVid
& 65.9 & 0.283 & 0.295 & 0.461 & 0.328 & 0.190
& 0.879 & 0.896 & 0.982 & 0.767 \\

\rowcolor{black!5}\multicolumn{11}{l}{\textit{\textbf{Memory / positional methods}}}\\
MemRoPE
& \underline{72.1} & \underline{0.640} & 0.578 & 0.559 & \underline{0.666} & 0.551
& 0.892 & 0.908 & \textbf{0.986} & 0.700 \\

\rowcolor{black!5}\multicolumn{11}{l}{\textit{\textbf{Spatiotemporal memory routing (ours)}}}\\
\rowcolor{claimblue}
\textbf{TetherMem}
& 71.1 & \textbf{0.780} & \textbf{0.769} & \underline{0.600} & \textbf{0.708}
& \textbf{1.289} & 0.855 & 0.895 & 0.973 & \textbf{0.867} \\
\bottomrule
\end{tabularx}
\endgroup
\caption{\textbf{Main comparison with human preference and VBench diagnostics.}
Human values are tie-aware Davidson EP on complete 30-second videos. VBench uses
270 evaluation videos: Subj./Back. use complete clips, while Smooth./Dyn. use a
uniform 8-second trim. Bold and underlined values mark best and second-best;
ID excludes the subject-free P05 prompt.}
\label{tab:main_sota}
\label{tab:expanded_vbench}
\end{table}

Table~\ref{tab:main_sota} shows that TetherMem has the highest overall and
progression estimates ($0.780$ and $0.769$), ahead of MemRoPE ($0.640$) and
LongLive-RAG ($0.608$), respectively. The identity and progression columns
expose the intended balance: Rolling Forcing has the highest identity estimate
($0.633$) but a progression EP of $0.429$, whereas TetherMem combines the
second-highest identity estimate ($0.600$) with the highest progression EP.
TetherMem also has the highest visual-integrity point estimate ($0.708$) and
nT ($1.289$). Img5 remains essentially unchanged from the backbone ($71.1$
versus $71.0$).

\subsection{Statistical Reliability and Prompt Robustness}
\label{sec:statistical_reliability}

\begin{table}[H]
\centering
\begingroup
\small
\setlength{\tabcolsep}{4.0pt}
\renewcommand{\arraystretch}{1.05}
\textbf{(a) Crossed-cluster uncertainty}\par\vspace{2pt}
\begin{tabular*}{\textwidth}{@{\extracolsep{\fill}}lclcc@{}}
\toprule
Criterion & TetherMem EP (95\% CI) & Strongest baseline & Baseline EP & Difference (95\% CI) \\
\midrule
Ovr.  & 0.780 [0.684, 0.868] & MemRoPE & 0.640 & 0.140 [0.004, 0.273] \\
Prog. & 0.769 [0.666, 0.861] & LongLive-RAG & 0.608 & 0.161 [0.076, 0.240] \\
ID    & 0.600 [0.504, 0.699] & Rolling Forcing & 0.633 & $-0.033$ [$-0.182$, 0.111] \\
L1    & 0.708 [0.625, 0.794] & MemRoPE & 0.666 & 0.042 [$-0.086$, 0.182] \\
\bottomrule
\end{tabular*}

\vspace{7pt}
\textbf{(b) Prompt-cluster sensitivity}\par\vspace{2pt}
\begin{tabular*}{\textwidth}{@{\extracolsep{\fill}}lccc@{}}
\toprule
Criterion & TetherMem EP & Strongest baseline EP & Difference (95\% CI) \\
\midrule
Ovr.  & 0.780 & 0.640 (MemRoPE) & 0.140 [$-0.017$, 0.290] \\
Prog. & 0.769 & 0.608 (LongLive-RAG) & 0.161 [0.074, 0.258] \\
\bottomrule
\end{tabular*}
\endgroup
\caption{\textbf{Statistical reliability across evaluation units and prompts.}
\textbf{(a)} Percentile intervals from 2,000 crossed annotator--prompt-seed
cluster bootstrap refits. \textbf{(b)} A stricter analysis clusters the three
seeds of each prompt before resampling; 1,995 of 2,000 refits are finite.}
\label{tab:statistical_robustness}
\label{tab:davidson_uncertainty}
\end{table}

Table~\ref{tab:statistical_robustness}(a) reports uncertainty against the
strongest baseline for each criterion. Under crossed annotator--prompt-seed
clustering, the overall margin is $0.140$ (95\% CI $[0.004,0.273]$) and the
progression margin is $0.161$ (95\% CI $[0.076,0.240]$); both intervals exclude
zero. Under the stricter prompt-level clustering in
Table~\ref{tab:statistical_robustness}(b), only the progression interval remains
separated from zero. ID and L1 remain comparable to their leading baselines.

\subsection{Ablations and Design Controls}
\label{sec:ablation_studies}

Table~\ref{tab:ablation} reports the routing-component and post-attention
design controls; the main comparisons are interpreted directly below the table.

\begin{table}[H]
\centering
\begingroup
\small
\setlength{\tabcolsep}{5.2pt}
\renewcommand{\arraystretch}{1.08}
\begin{tabularx}{\textwidth}{l*{6}{>{\centering\arraybackslash}X}}
\toprule
Variant & Ovr. & Prog. & ID & L1 & nT$\uparrow$ & Tail subj.$\uparrow$ \\
\midrule
\rowcolor{black!5}\multicolumn{7}{l}{\textbf{(a) Routing components: Full-Memory-centered comparison}}\\
Full Memory        & 0.426 & 0.391 & 0.472 & 0.457 & 0.980 & -- \\
Routing w/o Region & 0.461 & \underline{0.472} & 0.489 & \underline{0.505} & \underline{1.232} & -- \\
Routing w/o Age    & \underline{0.476} & 0.428 & \underline{0.505} & 0.490 & 1.083 & -- \\
\rowcolor{claimblue}
\textbf{TetherMem} & \textbf{0.638} & \textbf{0.709} & \textbf{0.534} & \textbf{0.548} & \textbf{1.458} & -- \\
\midrule
\rowcolor{black!5}\multicolumn{7}{l}{\textbf{(b) Post-attention design baseline}}\\
Full Memory        & \underline{0.513} & \underline{0.483} & \underline{0.634} & \underline{0.557} & 0.980 & 0.865 \\
Value Reweighting  & 0.174 & 0.303 & 0.108 & 0.230 & 1.405 & 0.859 \\
\rowcolor{claimblue}
\textbf{TetherMem} & \textbf{0.812} & \textbf{0.714} & \textbf{0.758} & \textbf{0.713} & \textbf{1.468} & \textbf{0.927} \\
\bottomrule
\end{tabularx}

\endgroup
\caption{\textbf{Routing-component and design ablations.}
Human columns are regularized, tie-aware Davidson EP fitted separately within
each panel. \textbf{(a)} Four-way routing comparison over 14 prompt--seed units.
\textbf{(b)} Three-way design comparison over seven prompts and two seeds; nT
and final-five-second subject consistency are averaged over 14 cells.}
\label{tab:ablation}
\label{tab:value_reweight_ep}
\label{tab:v2_ablation_auto}
\end{table}

Table~\ref{tab:ablation}(a) follows the mechanism from uniform history to the
complete router. Routing w/o Region and Routing w/o Age improve complementary
criteria; their combination leads every human dimension and nT. In
Table~\ref{tab:ablation}(b), Value Reweighting raises nT but lowers human EP and
tail-subject consistency, whereas normalized routing gives the strongest
progression--identity balance.

\subsection{Qualitative Trajectories}
\label{sec:more_analysis}

\begin{figure}[t]
\centering
\includegraphics[width=\linewidth]{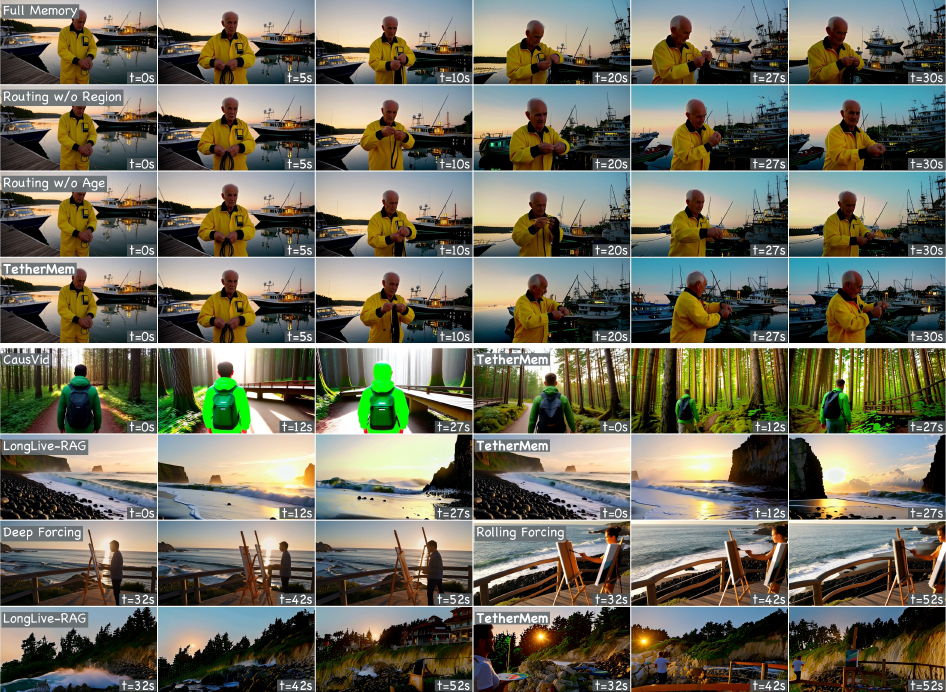}
\caption{\textbf{Compact memory-control and cross-method comparison.}
\textbf{Top four rows:} matched P01 trajectories at
0/5/10/20/27/30 seconds for Full Memory, the two single-factor routing
controls, and TetherMem. \textbf{Bottom four rows:} paired P10 and P05
trajectories at 0/12/27 seconds, followed by two independent P09 Painter Coast
comparisons at 32/42/52 seconds.
The extended 80-frame gallery with denser temporal sampling appears in
Appendix~Figure~\ref{fig:ablation_vis_extended}.}
\label{fig:ablation_vis}
\end{figure}

The matched P01 rows visualize the progression from uniform memory to complete
routing. Full Memory repeatedly reconstructs closely related harbor states;
regional routing protects subject-associated access, and recency routing
prevents the background from repeatedly returning to old scene states. Their
combination preserves the fisherman while allowing the harbor view to develop.
The lower rows retain the forest-hiker and subject-free beach comparisons,
followed by two independent Painter Coast comparisons. The final two rows place
Deep Forcing, Rolling Forcing, LongLive-RAG, and TetherMem on the same P09
prompt and time grid.

\begin{figure}[H]
\centering
\includegraphics[width=\linewidth]{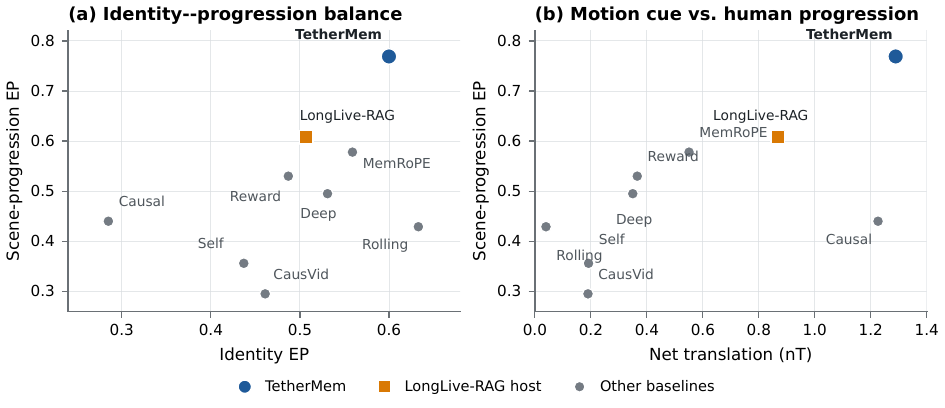}
\caption{\textbf{The evaluation landscape exposes two distinct gaps.}
\textbf{(a)} Identity EP versus scene-progression EP shows that strong identity
can coexist with weak progression. \textbf{(b)} nT versus human progression
shows that coherent translation is informative but incomplete: Causal Forcing
has high nT and low human progression. Values are the method-level
aggregates from Table~\ref{tab:main_sota}.}
\label{fig:evaluation_landscape}
\end{figure}

Figure~\ref{fig:evaluation_landscape} separates identity from progression and
shows why nT is complementary. Detailed VBench diagnostics appear in
Appendix~\ref{sec:vbench_diagnostics}.

\subsection{Realized Routing}
\label{sec:realized_routing}

\begin{figure}[H]
\centering
\includegraphics[width=\linewidth]{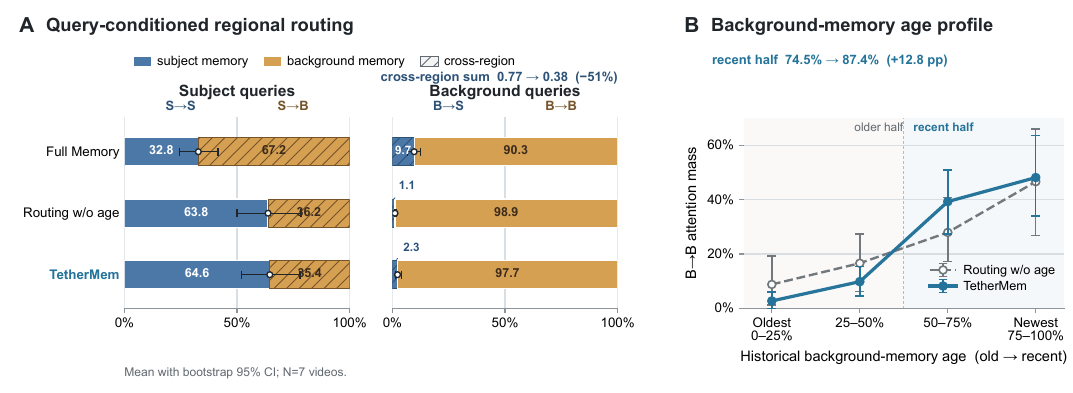}
\caption{\textbf{Realized routing behavior.}
\textbf{A} Regional routing reduces cross-region historical access; the two
routed variants have closely matched regional profiles.
\textbf{B} Within background memory, TetherMem redistributes attention toward
the recent half relative to Routing w/o Age. Points and bars show means with
video-level bootstrap 95\% CIs over seven videos.}
\label{fig:routing_evidence}
\end{figure}

Figure~\ref{fig:routing_evidence} examines realized post-softmax attention.
Subject-to-background attention falls from $67.2\%$ with Full Memory to
$36.2\%$ with regional routing and $35.4\%$ with TetherMem;
background-to-subject attention falls from $9.7\%$ to $1.1\%$ and $2.3\%$,
respectively. Within background memory, the recency prior shifts attention
away from the oldest half toward the most recent half. Together, the two panels
connect the designed regional and age priors to the attention actually realized
by the generator.

\begin{figure*}[p]
\centering
\includegraphics[width=\textwidth]{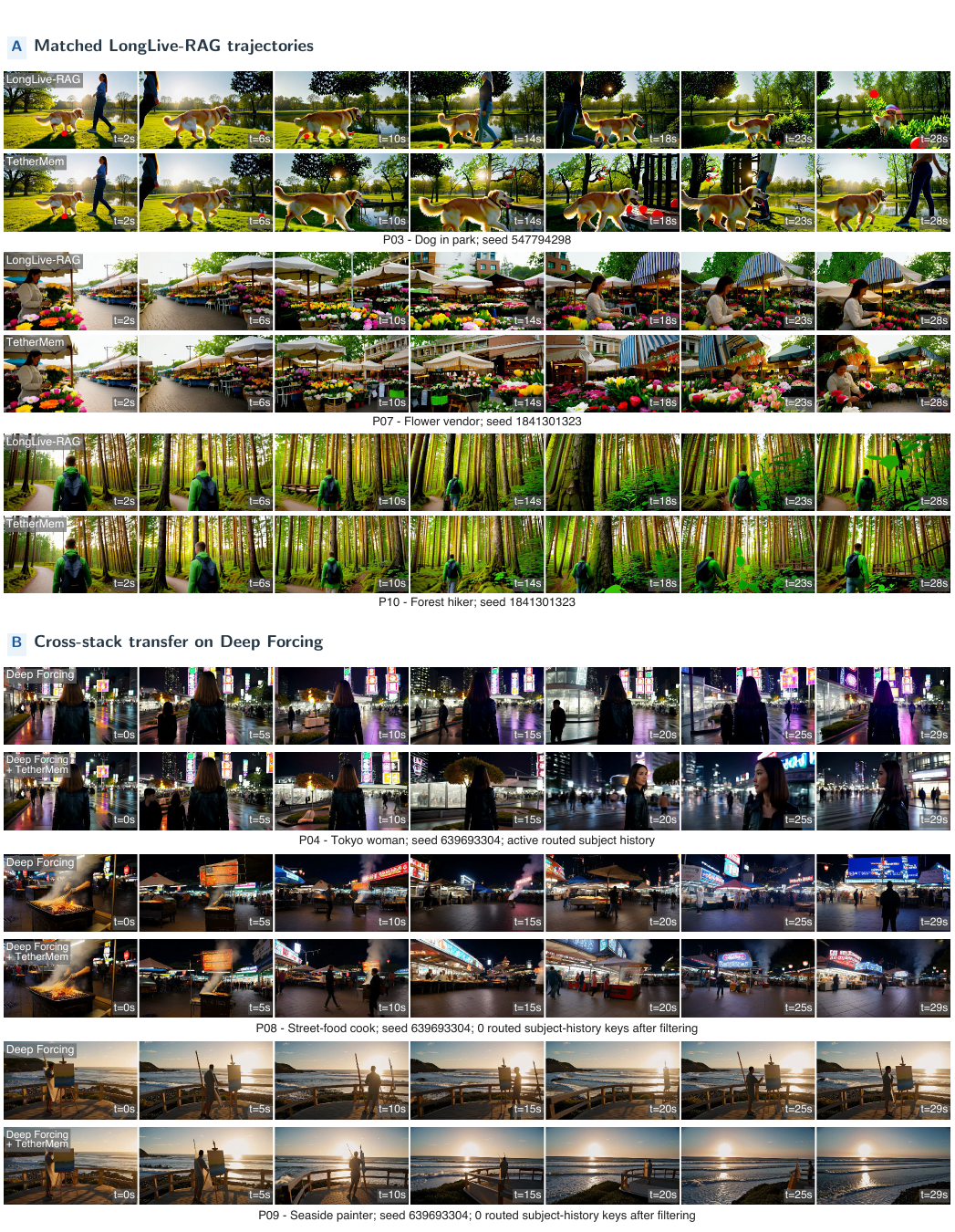}
\caption{\textbf{Long-horizon trajectories and cross-stack transfer.}
\textbf{A} Matched LongLive-RAG and TetherMem trajectories for P03, P07, and
P10 at 2--28 seconds. \textbf{B} Deep Forcing versus Deep Forcing+TetherMem
for P04, P08, and P09 at 0--29 seconds; the Wan2.1 foundation family is held
fixed while the autoregressive host stack changes.}
\label{fig:matched_trajectories_main}
\end{figure*}

\subsection{Transfer Across Autoregressive Hosts}
\label{sec:cross_stack_main}

Figure~\ref{fig:matched_trajectories_main}, panel A, follows the same LongLive-RAG host
pair through complete trajectories: LongLive-RAG often preserves a locally
coherent composition, whereas TetherMem maintains the central subject while
more of the scene enters the trajectory. Panel B applies the same router to a
Deep Forcing host within the Wan2.1 foundation family.

\begin{table}[H]
\centering
\begingroup
\small
\setlength{\tabcolsep}{5.2pt}
\renewcommand{\arraystretch}{1.05}
\begin{tabularx}{\textwidth}{l*{4}{>{\centering\arraybackslash}X}}
\toprule
& \multicolumn{4}{c}{\textbf{30s human EP}$\uparrow$} \\
\cmidrule(lr){2-5}
Method & Ovr. & Prog. & ID & L1 \\
\midrule
Deep Forcing & 0.404 & 0.360 & 0.470 & 0.439 \\
\rowcolor{claimblue}
\textbf{Deep Forcing + TetherMem} & \textbf{0.596} & \textbf{0.640} & \textbf{0.530} & \textbf{0.561} \\
\bottomrule
\end{tabularx}
\endgroup
\caption{\textbf{Blinded transfer to the Deep Forcing host.}
Thirty matched prompt--seed pairs receive three judgments each. The progression
difference is 0.281 with crossed-cluster 95\% CI [0.033, 0.522].}
\label{tab:cross_stack_main}
\end{table}

Across the complete 30-pair blind pool, Table~\ref{tab:cross_stack_main} raises
progression EP from $0.360$ to $0.640$; Overall, ID, and L1 also increase. Direct
counts, confidence intervals, and execution differences are reported in
Appendix~\ref{sec:deep_forcing_blind_audit}.

\subsection{Prior Robustness}
\label{sec:robustness_cost}

\begin{table}[H]
\centering
\begingroup
\small
\setlength{\tabcolsep}{5.0pt}
\renewcommand{\arraystretch}{1.05}
\begin{tabular*}{\textwidth}{@{\extracolsep{\fill}}lccc@{}}
\toprule
Prior & nT$\uparrow$ & Artifact veto$\downarrow$ & Tail subject$\uparrow$ \\
\midrule
Original SAM~2 & 1.490 & 0.158 & \textbf{0.934} \\
Low-res coarse & 1.463 & \textbf{0.107} & 0.932 \\
Bounding box   & \textbf{1.664} & 0.207 & 0.878 \\
\bottomrule
\end{tabular*}
\endgroup
\caption{\textbf{Sensitivity to subject-prior approximation.}
Automatic diagnostics over seven matched prompt-videos (one seed).
Artifact veto is the fraction flagged by the artifact classifier; Tail subject
is subject consistency in the final five seconds.}
\label{tab:robustness_cost}
\label{tab:prior_sensitivity}
\end{table}

The reference mask and a controlled-output re-extraction agree at
IoU $0.431$ over the complete rollout and $0.275$ in the late window. At
28 seconds, an independent human mask comparison measures median coverage $0.574$ and
raw IoU $0.202$ for the consumed reference prior across 21 subject-present
frames. Detailed temporal and human-mask results are in
Appendix~\ref{sec:prior_alignment_audit}.

Table~\ref{tab:robustness_cost} tests spatial approximation with checkpoint,
noise, reference rollout, and routing constants fixed. A coarse $8\times13$
prior remains close to the original on nT and tail-subject consistency, whereas
a bounding box increases the artifact-veto rate and reduces tail-subject
consistency. The result shows tolerance to boundary coarsening and sensitivity
to loose background inclusion.

\FloatBarrier

\section{Conclusion}
\label{sec:conclusion}

We identify memory-anchored scene under-progression, where uniform historical
access preserves local stability but suppresses prompted scene evolution.
TetherMem addresses this failure through normalized regional and recency
routing, retaining long-range subject evidence while directing background
queries toward recent scene states. On the evaluated 30-second suite, this
query-aware routing improves scene progression and overall preference while
preserving subject identity. Realized-attention measurements link the gains to
the regional and recency priors.
\FloatBarrier

\FloatBarrier
\bibliography{tethermem}

\clearpage
\raggedbottom
\appendix

\section{Additional Results and Details}
\label{sec:supp_overview}

This appendix provides additional qualitative results, the human-evaluation
protocol, prompt and baseline configurations, metric definitions,
implementation details, and ablations. Unless stated otherwise, experiments use
the Wan2.1-T2V-1.3B/LongLive-RAG stack, ten prompts, three seeds, and
approximately 30-second videos. Longer rollouts are included as qualitative
examples.

\section{Evaluation and Generation Setup}
\label{sec:setup_overview}

All long-video methods use matched prompts and seeds, 832$\times$480 output,
16 fps, and approximately 30-second duration. TetherMem and the Full-Memory
reference share the generator, checkpoints, denoising schedule, context,
attention sinks, and retrieval budget; they differ only in historical-memory
routing. Section~\ref{sec:prompts_protocol} lists the prompts and baseline
configurations.

For subject-containing prompts, the Full-Memory reference video is processed
by SAM~2 without frame-wise manual correction. The resulting frame-aligned
binary track is mapped to the $30\times52$ latent-token grid and supplies the
subject/scene roles used by the controlled rollout. The subject-free P05 prompt
uses a fixed center region as the default spatial partition. The complete
pipeline consists of the reference rollout, mask extraction, and controlled
generation.

\section{Qualitative Results}
\label{sec:visual_evidence}

\begin{figure*}[p]
  \centering
  \includegraphics[width=\textwidth]{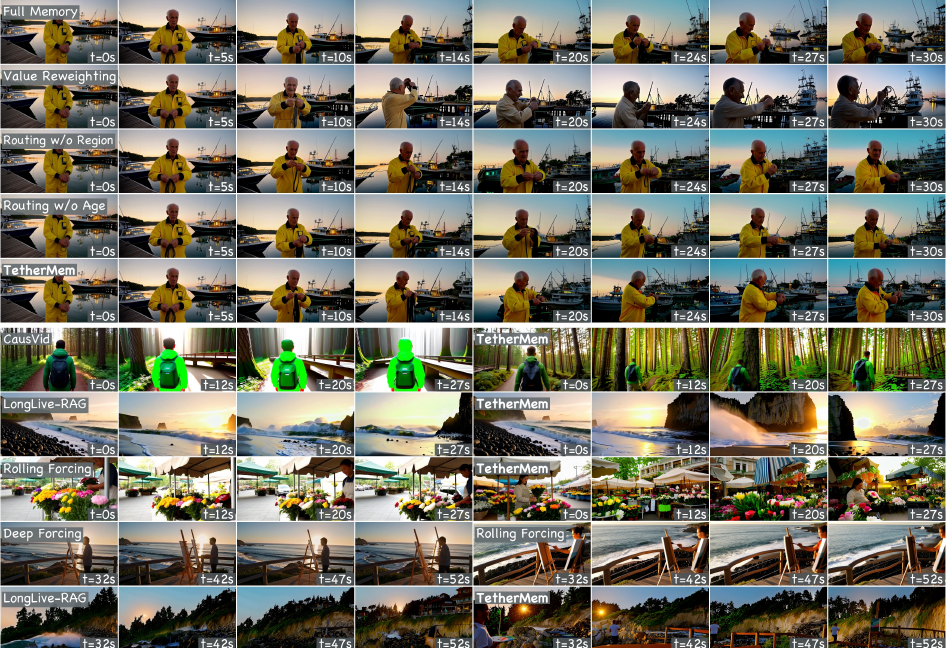}
  \caption{\textbf{Extended memory-control and cross-method gallery.}
  The top five rows show eight sampled times for Full Memory, Value
  Reweighting, Routing w/o Region, Routing w/o Age, and TetherMem on P01. The
  bottom rows compare CausVid/TetherMem on P10, LongLive-RAG/TetherMem on P05,
  Rolling Forcing/TetherMem on P07, and two P09 Painter Coast trajectories.}
\label{fig:ablation_vis_extended}
\end{figure*}

\begin{figure*}[p]
  \centering
  \includegraphics[width=\textwidth]{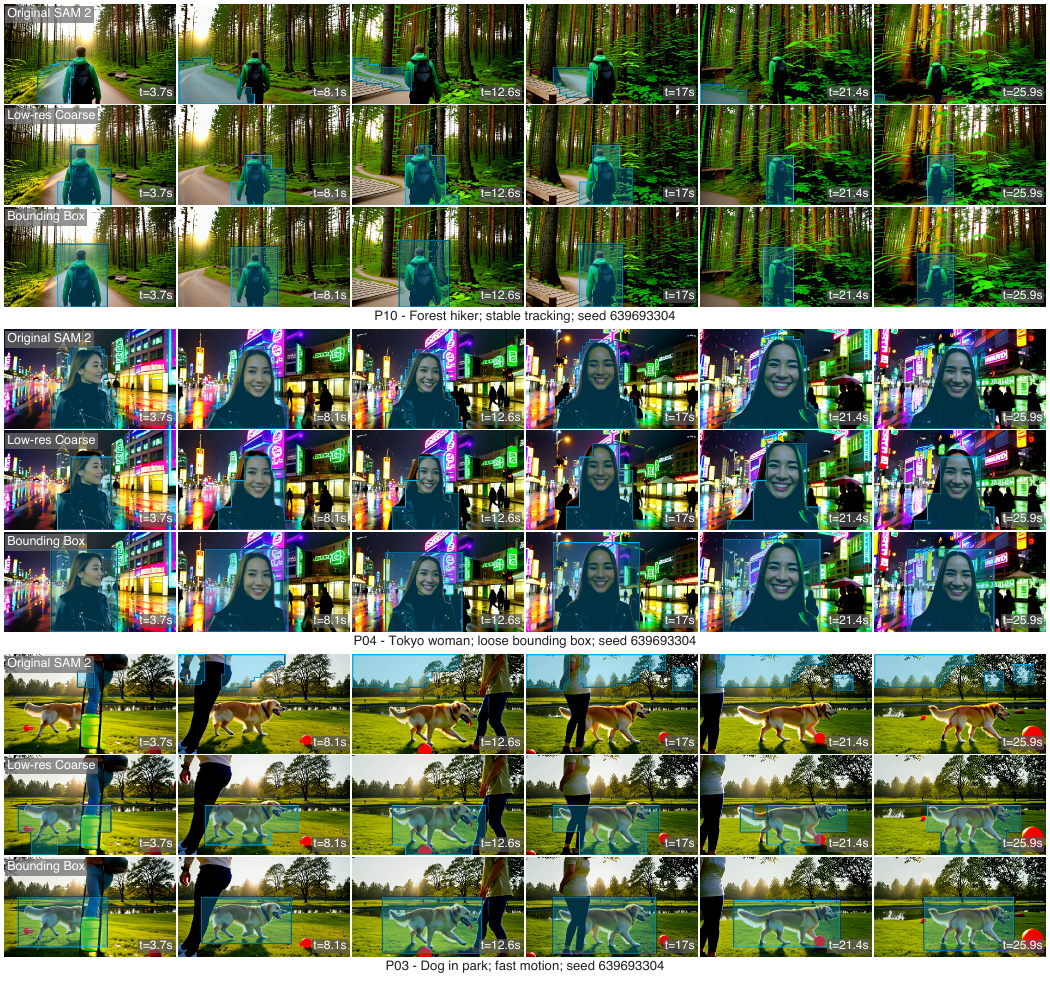}
  \caption{\textbf{Automatic subject priors at the routing grid.}
  Prompt blocks: P10, P04, and P03. Rows: original SAM~2, coarse, and
  bounding-box priors. Columns: six frames from 3.7 to 25.9 seconds. Cyan
  overlays show the binary $30\times52$ routing prior.}
\label{fig:supp_mask_prior}
\end{figure*}

Figures~\ref{fig:ablation_vis_extended} and~\ref{fig:supp_mask_prior} show
additional trajectories and the subject masks used at the routing grid.
Timestamps are embedded in the frames, and prompt--seed metadata appears below
each group. Figure~\ref{fig:matched_trajectories_main}, panel B, shows the same
router on a Deep Forcing host using the
\texttt{self\_forcing\_dmd.pt} EMA checkpoint. The quantitative comparison uses
all ten prompts and three seeds.

\subsection{Cross-Stack Human Evaluation}
\label{sec:deep_forcing_blind_audit}

The evaluation contains 30 matched Deep Forcing versus Deep Forcing + TetherMem
pairs from ten prompts and three seeds. Three blinded annotators judged each
pair, giving 90 judgments; six hidden repeats were excluded from the method
estimates. We use the same tie-aware Davidson--Bradley--Terry model as in the
main comparison and 2,000 crossed-cluster bootstrap refits over annotators and
prompt--seed units.

Table~\ref{tab:cross_stack_main} reports progression EP of 0.640 for the routed
host and 0.360 for Deep Forcing, a difference of 0.281 with 95\% CI
[0.033, 0.522]. Overall, ID, and L1 have higher point estimates, but their
difference intervals include zero. Direct W/T/L/U counts for the routed host
are 36/34/19/1 (Overall), 40/34/15/1 (Prog.), 25/39/20/6 (ID), and 29/43/18/0
(L1). All six hidden repeats agree on Overall, and 21 of 24 repeat-criterion
responses agree.
\FloatBarrier

\section{Human-Evaluation Protocol}
\label{sec:human_stats}

\subsection{Annotation Rubric and Procedure}

The rubric evaluates five dimensions. L1 covers technical defects and
late-stage collapse. L2 covers subject identity and continuity and is marked
N/A for the subject-free prompt. L3 measures spatial and scene development;
freezing, rollback, repetition, stationary jitter, and texture-only motion do
not count as progress. L4 measures progress toward the prompted action, event,
spatial relation, or viewpoint change. Overall asks which complete video is
preferred after considering quality, identity, progression, and long-horizon
viewing experience.

We fit L3 and L4 separately and average their EP scores only afterward. Thus,
the combined progression score reflects both spatial development and
prompt-directed semantic progress rather than motion alone.

Annotators viewed anonymous side-by-side videos with randomized left--right
order and selected ``left better,'' ``approximately equal,'' ``right better,''
or ``unsure'' for each criterion; L2 additionally allowed N/A. Hidden repeats
were interleaved with ordinary pairs and excluded from model fitting. A sample
task is shown in Figure~\ref{fig:annotation_interface_audit}.

\subsection{Tie-Aware Common Preference Scale}

The main comparison contains eight TetherMem--baseline edges and no direct
baseline--baseline comparisons. We use the Davidson tie extension of
Bradley--Terry~\cite{davidson1970ties}. For methods
$i,j$ with latent log-strengths $\theta_i,\theta_j$ and tie parameter $\nu>0$,
we model
\begin{align}
P(i>j)&=\frac{e^{\theta_i}}{D_{ij}},\\
P(i=j)&=\frac{2\nu e^{(\theta_i+\theta_j)/2}}{D_{ij}},\\
D_{ij}&=e^{\theta_i}+e^{\theta_j}
       +2\nu e^{(\theta_i+\theta_j)/2}.
\end{align}
We fix $\theta_{\text{TetherMem}}=0$ for identifiability and fit one tie
parameter per criterion by maximum likelihood.

We convert the fitted model to a common, interpretable expected-preference (EP)
score
\begin{equation}
S_i=\frac{1}{K-1}\sum_{j\ne i}
\left[P(i>j)+\tfrac{1}{2}P(i=j)\right].
\end{equation}
It is the expected score against a uniformly sampled opponent when
win/tie/loss receive 1/0.5/0. Every row therefore uses the same opponent
distribution. Baseline--baseline probabilities are model predictions.

\subsection{Data and Uncertainty}

The analysis contains 2,400 judgments from 10 annotators who completed the full
assignment. One partially completed assignment is excluded at the annotator
level. Ties are retained in the likelihood; unsure and N/A responses are
excluded; hidden repeats are used only for quality control. L3 and L4 are fit
independently, and Prog. is their average inside each resample.

To account for shared annotators and prompt--seed units, we use 2,000 crossed
cluster bootstrap refits~\cite{wu2022diagnostic}. Annotators and the 30
prompt--seed units are resampled independently. Table~\ref{tab:davidson_uncertainty}(a)
reports comparisons with the strongest baseline for each criterion.

\subsection{Direct Observations and Sensitivity}

Table~\ref{tab:direct_outcomes}(a) gives direct Overall outcomes on each
TetherMem--baseline edge. The half-tie score is
$(W+0.5T)/(W+T+L)$; the decided-only rate is $W/(W+L)$. They are reported
separately from the common-scale EP scores in Table~\ref{tab:main_sota}.

Table~\ref{tab:direct_outcomes}(b) reports the two directly observed
progression criteria. These counts use TetherMem's perspective; the common-scale
EP scores use the fitted graph. In the backbone-controlled
LongLive-RAG comparison, the half-tie scores are $0.687$ for L3 and $0.654$
for L4.

\begin{table}[t]
\centering
\begingroup
\begin{minipage}[t]{0.43\textwidth}
\vspace{0pt}
\centering
\textbf{(a) Overall preference}\par\vspace{2pt}
{\fontsize{6.2}{7.1}\selectfont
\setlength{\tabcolsep}{1.3pt}
\renewcommand{\arraystretch}{1.04}
\begin{tabular}{@{}lrrrrrr@{}}
\toprule
Baseline & W & T & L & U & Half & Dec. \\
\midrule
CausVid         & 254 & 8  & 27 & 11 & 0.893 & 0.904 \\
Causal Forcing  & 259 & 22 & 12 & 7  & 0.922 & 0.956 \\
Deep Forcing    & 207 & 25 & 59 & 9  & 0.754 & 0.778 \\
LongLive-RAG    & 157 & 92 & 49 & 2  & 0.681 & 0.762 \\
MemRoPE         & 173 & 41 & 80 & 6  & 0.658 & 0.684 \\
Reward Forcing  & 193 & 29 & 67 & 11 & 0.718 & 0.742 \\
Rolling Forcing & 206 & 24 & 67 & 3  & 0.734 & 0.755 \\
Self Forcing    & 243 & 24 & 24 & 9  & 0.876 & 0.910 \\
\bottomrule
\end{tabular}}
\end{minipage}\hfill
\begin{minipage}[t]{0.55\textwidth}
\vspace{0pt}
\centering
\textbf{(b) Scene progression}\par\vspace{2pt}
{\fontsize{6.2}{7.1}\selectfont
\setlength{\tabcolsep}{1.4pt}
\renewcommand{\arraystretch}{1.04}
\begin{tabular}{@{}lrrrrrrrr@{}}
\toprule
& \multicolumn{4}{c}{L3}
& \multicolumn{4}{c}{L4} \\
\cmidrule(lr){2-5}\cmidrule(lr){6-9}
Baseline & W & T & L & U & W & T & L & U \\
\midrule
CausVid         & 247 & 25  & 17 & 11 & 221 & 39  & 28 & 12 \\
Causal Forcing  & 188 & 95  & 11 & 6  & 178 & 105 & 11 & 6  \\
Deep Forcing    & 200 & 55  & 36 & 9  & 177 & 74  & 41 & 8  \\
LongLive-RAG    & 137 & 138 & 25 & 0  & 123 & 144 & 31 & 2  \\
MemRoPE         & 178 & 66  & 52 & 4  & 163 & 74  & 58 & 5  \\
Reward Forcing  & 198 & 48  & 46 & 8  & 175 & 61  & 56 & 8  \\
Rolling Forcing & 231 & 28  & 38 & 3  & 205 & 47  & 44 & 4  \\
Self Forcing    & 231 & 47  & 16 & 6  & 211 & 56  & 28 & 5  \\
\bottomrule
\end{tabular}}
\end{minipage}
\endgroup
\caption{\textbf{Direct observed outcomes from TetherMem's perspective.}
\textbf{(a)} Overall preference, including the half-tie score (Half) and
decided-only rate (Dec.). \textbf{(b)} Spatial and scene progression (L3) and
prompt-directed progression (L4). W/T/L/U denote win/tie/loss/unsure. Prog. in
the main text averages the separately fitted L3 and L4 EP scores rather than
pooling these counts.}
\label{tab:direct_outcomes}
\label{tab:direct_wtl}
\label{tab:direct_progress_wtl}
\end{table}

Leave-one-annotator-out TetherMem score ranges are [0.761, 0.808] (Ovr.),
[0.748, 0.800] (Prog.), [0.570, 0.618] (ID), and [0.683, 0.727] (L1).
Hidden-repeat agreement is 90/120 (0.750) for Overall and 433/600 (0.722)
across the five dimensions.

For prompt-level sensitivity, we group the three seeds of each prompt into one
cluster before resampling. Of 2,000 refits, 1,995 remain finite. The
TetherMem--strongest-baseline differences are $0.140$ [${-0.017}$, $0.290$]
for Overall and $0.161$ [$0.074$, $0.258$] for Prog. The progression interval
excludes zero; the Overall interval does not.

\section{Detailed Evaluation Suite and Executed Configurations}
\label{sec:prompts_protocol}

\subsection{Prompt Suite and Prompt-Level Robustness}

The evaluation uses three prompt groups: stable-scene controls (P01--P03),
subject-centered progression (P04 and P06--P10), and subject-free progression
(P05). The three seeds are 639693304, 1841301323, and 547794298.

\begin{description}
\small
\item[P01.] A tight half-body shot of an elderly fisherman in a yellow slicker standing on a wooden harbor pier at dusk, coiling a rope on his forearm. Behind him, fishing boats bob in the still harbor and warm harbor lights begin to glow. The camera holds steady at his eye level; the pier, boats, and harbor remain stable as he works.
\item[P02.] A medium-close shot of a Buddhist monk in saffron robes standing still beneath a wooden temple gate at dawn, mist drifting past the pillars. A few bronze bells hang above him. The camera holds steady; the gate, bells, and misty valley behind remain stable as the monk breathes and the mist slowly shifts.
\item[P03.] A golden retriever trots across a sunlit park lawn toward its owner, a red rubber ball in the grass nearby, oak trees and a small pond in the background. The camera holds steady; the trees, pond, and park remain stable as the dog crosses the lawn.
\item[P04.] A stylish young woman in a black leather jacket walks along a rain-slicked Tokyo street at night, surrounded by glowing neon signs and passing pedestrians; colorful reflections shimmer on the wet pavement. Medium shot, camera slowly tracking backward in front of her.
\item[P05.] Waves roll onto an empty rocky beach as the tide slowly rises, under a sky shifting from pale dawn to warm sunrise; sea mist drifts past dark headland cliffs. Wide shot, static camera.
\item[P06.] A medium shot of a skier in a bright red jacket gliding downhill along a snowy mountain slope, carving smooth turns while keeping a steady posture. Pine trees line the slope, and distant mountain peaks gradually come into view as the camera tracks backward in front of the skier. Snow sprays lightly from the skis, while the alpine landscape opens up behind. Continuous motion, no abrupt cuts.
\item[P07.] A medium shot of a flower vendor arranging bouquets at an outdoor morning market, wearing a light apron and standing behind a stall filled with colorful flowers. The camera slowly tracks sideways, gradually revealing neighboring stalls, hanging awnings, and more of the bustling market street behind. The vendor remains the visual focus while the market scene naturally unfolds. No abrupt cuts, only smooth continuous camera movement.
\item[P08.] A medium shot of a street-food cook standing at a night market stall, turning skewers over a glowing grill as warm smoke rises into the air. The camera slowly tracks backward, gradually revealing more of the neon-lit market street, nearby lanterns, and passing customers in the background. The cook stays clearly visible in the foreground while the night-market scene continues to open up. No abrupt cuts, only smooth continuous camera movement.
\item[P09.] A medium shot of a painter standing beside an easel on a seaside boardwalk at sunset, brushing color onto a canvas while the ocean breeze moves the edge of the painting cloth. The camera slowly arcs around the painter, gradually revealing more of the rocky shoreline, the wooden railing, and distant waves crashing below. The painter remains clearly visible as the coastal scene opens up behind. No abrupt cuts, only smooth continuous camera movement.
\item[P10.] A medium shot of a hiker wearing a green jacket and carrying a small backpack, walking steadily along a forest trail. The camera slowly tracks backward in front of the hiker, gradually revealing more of the trail behind, tall trees, filtered sunlight, and a small wooden footbridge further down the path. The hiker remains clearly visible while the forest scene progressively unfolds. No abrupt cuts, only smooth continuous camera movement.
\end{description}

For the TetherMem--LongLive-RAG comparison, direct scores use
$(W+0.5T)/(W+T+L)$ from TetherMem's perspective; Prog. is the arithmetic mean
of the L3 and L4 direct scores. Overall and Prog. exceed $0.5$ on all ten
prompts.

Pooling within the three groups gives Overall/Prog. scores of
$0.756/0.767$ on the stable controls, $0.867/0.822$ on the subject-free prompt,
and $0.612/0.597$ on the six subject-centered progression prompts. On the
stable controls, direct Overall, L1, and ID scores are respectively
$0.756$, $0.644$, and $0.644$.

When each prompt is removed in turn, pooled direct Overall and Prog.
scores remain in $[0.660,0.694]$ and $[0.652,0.681]$. Re-fitting the complete
Davidson model after each deletion gives TetherMem Overall EP in
$[0.766,0.793]$ and Prog. EP in $[0.760,0.780]$; TetherMem remains rank one for
both criteria in all ten deletions.

\subsection{Generation Protocol}

All long-video methods use matched prompts, seeds, 832$\times$480 output, 16
fps, and approximately 30-second duration. The TetherMem/LongLive-RAG
configuration generates 120 latent frames (474 decoded frames) in blocks of
three with denoising timesteps $[1000,750,500,250]$. It uses local attention
size 12, retrieved-memory size 6, sink size 1, excludes the five most recent
frames from retrieval, and applies top-$k$ retrieval. TetherMem changes the
historical-memory routing while retaining this generator and rollout
configuration.

\subsection{Executed Baseline Configurations}

Tables~\ref{tab:baseline_configs} and~\ref{tab:output_adaptation} list the
settings recorded in the launch commands, configuration files, and generation
logs. Exact commit identifiers are unavailable. All methods use bfloat16 with
cuDNN disabled and conditional-only inference; each method retains its native
sampler and cache or memory settings.

\begin{table*}[t]
\centering
\begingroup
\scriptsize
\setlength{\tabcolsep}{3.0pt}
\renewcommand{\arraystretch}{1.08}
\begin{tabular}{@{}>{\raggedright\arraybackslash}p{1.9cm}
>{\raggedright\arraybackslash}p{2.9cm}
>{\raggedright\arraybackslash}p{3.1cm}
>{\raggedright\arraybackslash}p{3.6cm}
>{\raggedright\arraybackslash}p{2.8cm}@{}}
\toprule
Method & Checkpoint & Sampling & Chunk/context/memory & Method-specific settings \\
\midrule
LongLive-RAG
& \texttt{causal\_forcing.pt} + \texttt{ae\_latent\_mem.pt}
& 4 warped steps: 1000/750/500/250
& block 3; local 12; sink 1; memory 6; exclude-recent 5
& AE compression; top-$k$ retrieval \\
Self Forcing
& \texttt{self\_forcing\_dmd.pt}
& 4 warped steps: 1000/750/500/250; EMA
& block 3; full KV attention
& timestep shift 5.0 \\
Rolling Forcing
& \shortstack[l]{\texttt{rolling\_forcing\_}\\\texttt{dmd.pt}}
& 5 warped steps: 1000/800/600/400/200; EMA
& block 3; rolling-window length 5
& window advances one block \\
Deep Forcing
& \texttt{self\_forcing\_dmd.pt}
& 4 warped steps: 1000/750/500/250; EMA
& block 3; local 21; sink 14
& Budget 16 / Recent 4 supplied by the launch command \\
Causal Forcing
& \texttt{causal\_forcing.pt}
& 4 warped steps: 1000/750/500/250
& block 3; local 12; sink 1; retrieval disabled
& executed through the LongLive-RAG code path without latent memory \\
Reward Forcing
& \texttt{rewardforcing.pt}
& 4 warped steps: 1000/750/500/250; EMA
& block 3; local 9; sink 3
& timestep shift 5.0 \\
MemRoPE
& \texttt{self\_forcing\_dmd.pt}
& 4 warped steps: 1000/750/500/250; EMA
& block 3; local 12; sink 3; recent 4
& block-RoPE; EMA compression ($0.01/0.1$ long/short) \\
CausVid
& released autoregressive checkpoint
& 3 unwarped steps: 1000/757/522; shift 8.0
& block 3; seven 21-latent-frame rollouts; 3-frame overlap
& environment-variable seed; rollout concatenation and trim \\
TetherMem
& same two checkpoints as LongLive-RAG
& same 4 warped steps as LongLive-RAG
& same block, context, sink, memory, and retrieval settings
& region/age routing; $\alpha=0.25$, $\rho_{\min}=0.05$, anchor 1.0 \\
\bottomrule
\end{tabular}
\endgroup
\caption{\textbf{Executed inference configurations.} Settings correspond to
the runs reported in Table~\ref{tab:main_sota}.}
\label{tab:baseline_configs}
\end{table*}

All methods receive the same prompts and integer seeds, although implementation
differences mean that the resulting noise tensors are not identical across
codebases. The 270 videos are 832$\times$480 at 16 fps, with durations between
29.625 and 30.000 seconds. Table~\ref{tab:output_adaptation} lists the remaining
method-dependent differences.

\begin{table*}[t]
\centering
\begingroup
\scriptsize
\setlength{\tabcolsep}{4.0pt}
\renewcommand{\arraystretch}{1.06}
\begin{tabular}{@{}>{\raggedright\arraybackslash}p{5.0cm}cc
>{\raggedright\arraybackslash}p{7.2cm}@{}}
\toprule
Method group & Frames & Duration & 30-second adaptation \\
\midrule
LongLive-RAG, Causal Forcing, TetherMem
& 474 & 29.625 s & 120 latent frames; native single-pass causal rollout \\
Self Forcing, Rolling Forcing, Deep Forcing, Reward Forcing; MemRoPE
& 477 & 29.813 s & 120 latent frames; native single-pass or rolling KV rollout \\
CausVid
& 480 & 30.000 s & seven overlapping rollouts yield 504 frames, then trim and H.264 re-encode (CRF 18) \\
\bottomrule
\end{tabular}
\endgroup
\caption{\textbf{Final output adaptation.} No method uses looping,
frame-rate conversion, or spatial rescaling. CausVid alone applies the listed
trim and re-encode.}
\label{tab:output_adaptation}
\end{table*}

\section{Automatic Metrics and VBench Diagnostics}
\label{sec:auto_metric_definitions}

\subsection{Metric Definitions}

\textbf{Img5.} We run the VBench Imaging Quality dimension on the first five
seconds of each video and report $100$ times its per-video score. Values are
first computed per video and then averaged across prompt--seed cells.

\textbf{nT.} This is the evaluator's \texttt{net\_translation} diagnostic. We
decode the complete video, convert each frame to grayscale, and resize it to
width 200 while preserving aspect ratio. Dense flow is evaluated every two
frames ($\mathrm{stride}=2$) with OpenCV Farneback flow: pyramid scale 0.5,
three pyramid levels, window size 15, three iterations, polynomial neighborhood
5, polynomial sigma 1.2, and flags 0. Let
$\mathbf{f}_i(x)=(u_i(x),v_i(x))$ denote the dense flow for sampled transition
$i$, and let
\begin{equation}
\bar{\mathbf{f}}_i=\frac{1}{|\Omega|}\sum_{x\in\Omega}\mathbf{f}_i(x)
=(\mu_i,\nu_i)
\end{equation}
be its spatially averaged flow vector. For $N$ sampled transitions, we compute
\begin{equation}
\mathrm{nT}=\left\lVert\frac{1}{N}\sum_{i=1}^{N}
\bar{\mathbf{f}}_i\right\rVert_2
=\sqrt{\left(\frac{1}{N}\sum_i\mu_i\right)^2+
       \left(\frac{1}{N}\sum_i\nu_i\right)^2}.
\label{eq:nt_definition}
\end{equation}
Flow vectors are averaged over time before taking the norm. Opposing directions
therefore cancel, while sustained directional translation yields higher nT.

nT is computed per video and then averaged over prompt--seed cells. Its units
are pixels per sampled transition after resizing to width 200. It measures net
directional motion and does not encode subject identity or prompt fulfillment.

\textbf{Subject-consistency diagnostics.} Subj5 and Tail subj. apply the same
center-feature subject-consistency proxy to the first and final five seconds,
respectively. They are computed per video and then averaged over the relevant
prompt--seed cells.

\textbf{Artifact diagnostic.} Art. is a unit-interval artifact-veto score,
where lower is better. It is used only in the subject-prior sensitivity study.

\subsection{VBench Diagnostics on the Human-Evaluation Pool}
\label{sec:vbench_diagnostics}

Table~\ref{tab:expanded_vbench} reports four VBench
diagnostics~\cite{huang2023vbench} on the 270 videos in the main evaluation.
Subject and background consistency use the native videos; motion smoothness
and dynamic degree use a uniform 8-second trim.

At the method level ($N=9$), Spearman correlations with human progression EP
are $-0.217$ for subject consistency, $-0.100$ for background consistency,
$-0.171$ for motion smoothness, and $0.366$ for dynamic degree. Rolling Forcing
has the highest subject and background consistency but progression EP of
$0.429$; Causal Forcing ties for the highest dynamic degree but has progression
EP of $0.440$. TetherMem has the highest human progression EP without leading
the consistency metrics.

\FloatBarrier
\section{Implementation, Cost, and Prior Robustness}
\label{sec:implementation_details}

\subsection{Routing Configuration}

The routing configuration uses a subject anchor prior of one and a target
average spatial prior $\alpha=0.25$. Let $r_n$ be the fraction of
subject tokens in the current query frame. The implementation computes
\begin{align}
\bar{\gamma}_n
&=
\begin{cases}
(\alpha-r_n)/(1-r_n), & 1-r_n>10^{-6},\\
\alpha, & \text{otherwise},
\end{cases}\\
\gamma_n
&=\max\!\left(10^{-9},
  \min\!\left(1,\max(0,\bar{\gamma}_n)\right)\right).
\end{align}
The equation sets the area-weighted prior to $\alpha$ whenever feasible;
realized attention also depends on the query--key logits. We use
$\alpha=0.25$ and $\rho_{\min}=0.05$ for all evaluation runs. When
$r_n\geq\alpha$, the unclipped $\bar\gamma_n$ is non-positive and $\gamma_n$
therefore saturates at the numerical floor.

For age routing, let $p_i$ be the source frame's position in the memory pool,
where zero is the oldest position, and let $N_{\mathrm{pool}}$ be the current
pool length. The active absolute-recency configuration is
\begin{align}
A_{\max}&=\max\!\left(1,\min(N_{\mathrm{pool}},120)\right),\\
\tau_i&=p_i/A_{\max},&
\rho_i&=\max(\tau_i,0.05).
\end{align}
The two cross-region directions share $\gamma_n$, whereas the age prior applies
only when a background query reads a background memory key.

\begin{table}[t]
\centering
\begingroup
\scriptsize
\setlength{\tabcolsep}{4.0pt}
\renewcommand{\arraystretch}{1.04}
\begin{tabular}{@{}lll@{}}
\toprule
Query & Historical key & Prior \\
\midrule
Subject    & Subject    & $1$ \\
Subject    & Background & $\gamma_n$ \\
Background & Subject    & $\gamma_n$ \\
Background & Background & $\rho_i$ \\
Any        & Local/sink & $1$ \\
\bottomrule
\end{tabular}
\endgroup
\caption{\textbf{Routing priors.} Priors are converted to additive
logit biases before softmax.}
\label{tab:routing_configuration}
\end{table}

All priors enter scaled dot-product attention as $\log\pi_n(q,i)$ before
softmax. The implementation does not scale Q, K, V, or the attention output.
It splits subject and background queries, evaluates two attention calls against
the same complete key--value context, and scatters their outputs back without
an additional gate. The same routing is applied to all heads, denoising steps,
and 30 causal self-attention modules; the bias is shared across heads.

\subsection{Offline Subject Priors}

We generate a Full-Memory reference rollout and track one subject with the
SAM~2 Hiera-Large checkpoint using a box prompt~\cite{ravi2024sam2}. Masks are
binarized, resized to the $30\times52$ latent-token grid, and dilated by a
$5\times5$ element; dilation is reduced when the mask would exceed $0.25$ of
the grid. Current queries use the reference mask at the current time, and
historical keys use the mask at their source time. We denote this tensor by
$M^{\mathrm{ref}}_{\mathrm{used}}$.

The implementation tracks one subject. Missing masks restore the original
attention, empty frames use the default spatial partition, and a lost track
reuses the last valid mask. P05 uses the same default partition because the
prompt has no designated subject. No frame-wise manual correction is applied.
Section~\ref{sec:prior_alignment_audit} measures how the reference masks align
with the controlled rollout over time.

\begin{figure*}[!t]
  \centering
  \includegraphics[width=0.95\textwidth]{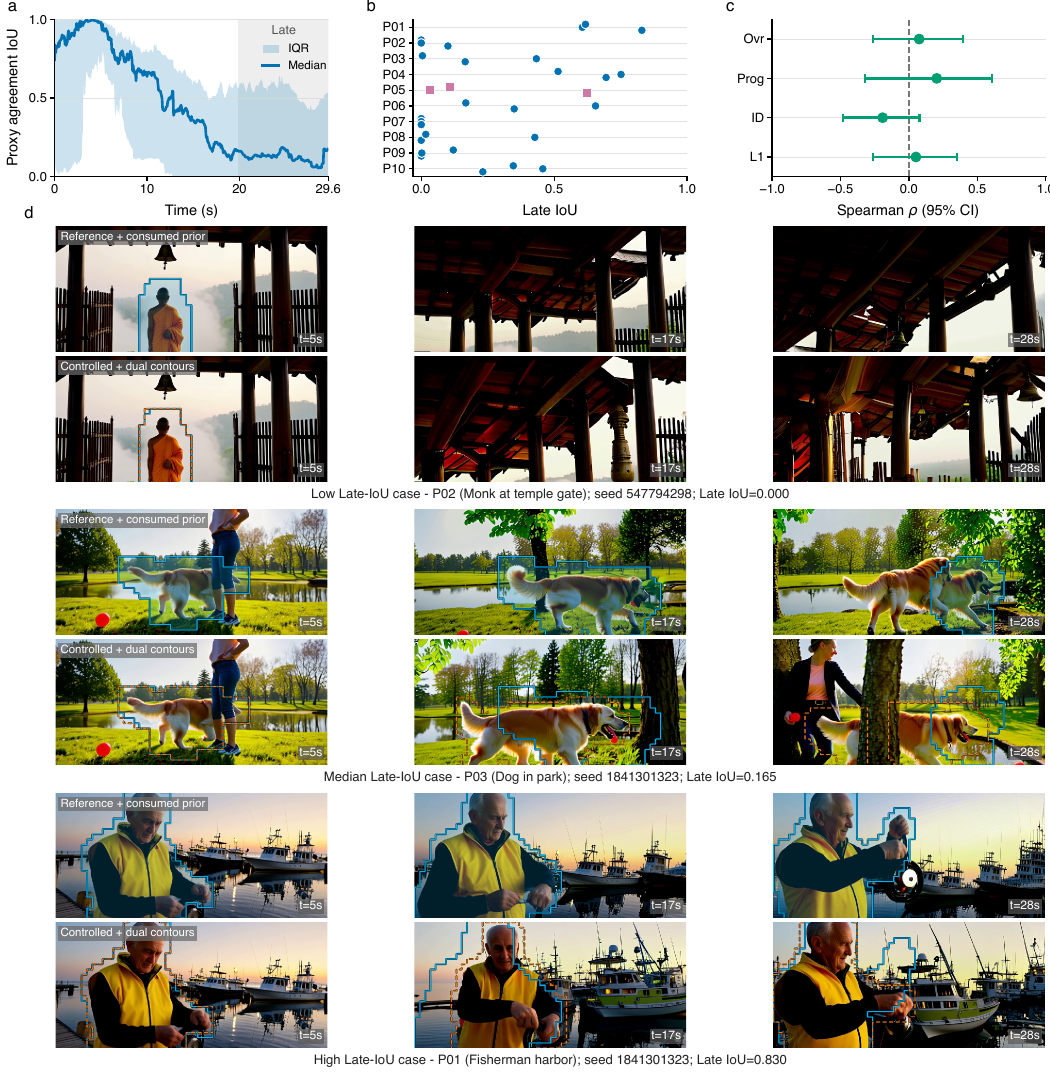}
  \caption{\textbf{Reference-prior drift across the 30 evaluation videos.}
  (a) Median and IQR over time. (b) Late IoU by prompt and seed.
  (c) Correlations with human scores. (d) Low, median, and high Late-IoU
  examples at 5, 17, and 28 seconds; cyan is the reference prior and orange is
  the controlled-output re-extraction.}
\label{fig:supp_prior_alignment}
\end{figure*}

\begin{figure*}[!t]
  \centering
  \includegraphics[width=0.94\textwidth]{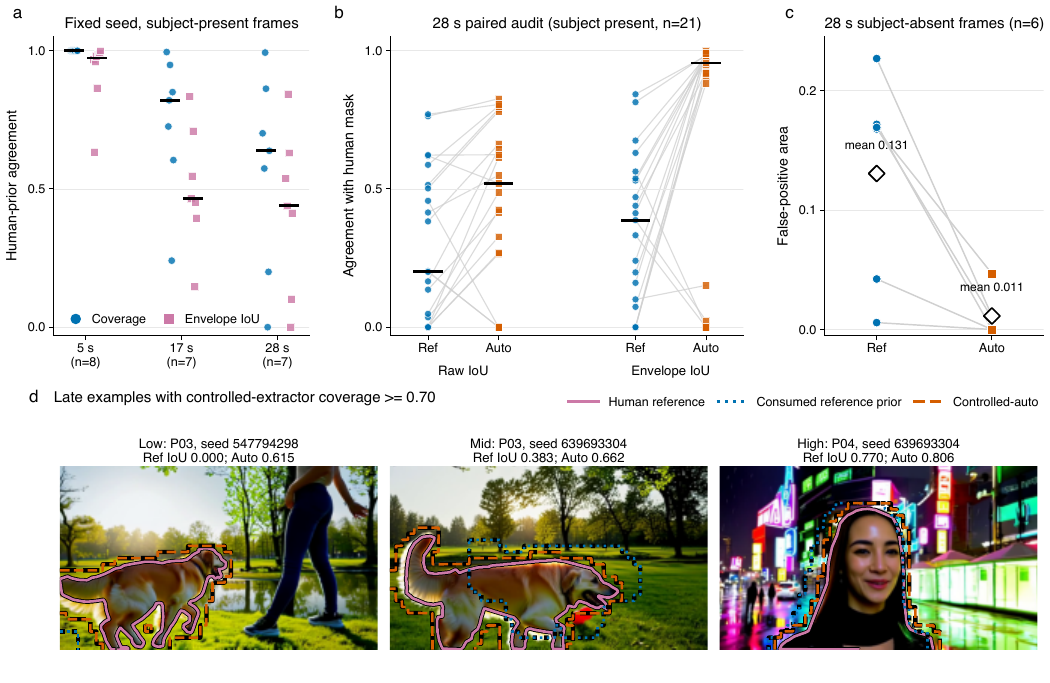}
  \caption{\textbf{Human mask comparison for the reference prior and controlled output.}
  (a) Agreement at 5, 17, and 28 seconds. (b) Paired comparison on 21
  subject-present frames at 28 seconds. (c) False-positive area on six
  target-absent frames. (d) Representative low, median, and high agreement
  cases. Contours: human mask (purple), reference prior (blue), and controlled
  re-extraction (orange).}
\label{fig:supp_human_semantic_prior_audit}
\end{figure*}

\clearpage

\subsection{Reference-Prior Drift}
\label{sec:prior_alignment_audit}

Because the reference and controlled trajectories can diverge, their subject
masks may become misaligned. We rerun YOLO--SAM~2 on all 30 controlled videos
(14,220 frames) to obtain $M^{\mathrm{ctrl}}_{\mathrm{auto}}$ and compare it
with $M^{\mathrm{ref}}_{\mathrm{used}}$. We call their IoU proxy agreement.

Proxy agreement is 0.431 with prompt-cluster 95\% CI $[0.306,0.564]$ over the
full rollout and 0.275 $[0.140,0.433]$ after 20 seconds. Subject coverage falls
from 0.583 overall to 0.424 in the late window, showing that the reference mask
becomes less aligned as the trajectories diverge.

\paragraph{Human mask comparison.}
One independent annotator marked the visible subject in 45 controlled-output
frames from P01--P04 and P06--P10 at 5, 17, and 28 seconds. P05 is excluded
because it has no designated subject. The annotator saw the RGB frame and
prompt but not the method name or automatic masks. This comparison uses one
annotator.

At 28 seconds, the target is present in 21 of 27 frames. Table~\ref{tab:human_semantic_prior_audit}
reports these 21 frames; false-positive area is measured on the six
target-absent frames. The controlled-output masks align more closely with the
human masks than the reference prior.

\begin{table}[t]
\centering
\begingroup
\scriptsize
\setlength{\tabcolsep}{3.3pt}
\renewcommand{\arraystretch}{1.04}
\begin{tabular*}{0.68\linewidth}{@{\extracolsep{\fill}}lcc@{}}
\toprule
28-s metric (subject present, $n=21$) & Ref. prior & Ctrl.-auto \\
\midrule
Subject coverage $\uparrow$ & 0.574 & 1.000 \\
Prior precision $\uparrow$ & 0.238 & 0.550 \\
Raw semantic IoU $\uparrow$ & 0.202 & 0.520 \\
Route-envelope IoU $\uparrow$ & 0.387 & 0.956 \\
Normalized centroid distance $\downarrow$ & 0.138 & 0.038 \\
\bottomrule
\end{tabular*}
\endgroup
\caption{\textbf{Mask alignment at 28 seconds.} Entries are medians over
the 21 subject-present frames. ``Ref. prior'' is the consumed
$M^{\mathrm{ref}}_{\mathrm{used}}$; ``Ctrl.-auto'' is the controlled-output
re-extraction.}
\label{tab:human_semantic_prior_audit}
\end{table}

The controlled-minus-reference difference is $+0.161$ with 95\% interval
$[0.001,0.273]$ for raw semantic IoU and $+0.389$ $[0.058,0.553]$ for
route-envelope IoU. Mean false-positive area on target-absent frames is 0.131
for the reference prior and 0.011 for the controlled-output mask. At 5, 17, and
28 seconds, median reference-prior coverage is 1.000, 0.821, and 0.638,
respectively. Prompt-cluster correlations between mask agreement and the four
human scores all have intervals that include zero
(Figure~\ref{fig:supp_prior_alignment}c).

The reference mask is well aligned early and less precise late in the rollout.

\FloatBarrier

\subsection{Runtime Accounting}

We profile seven videos at seed 639693304 on one NVIDIA H200. Timings are
CUDA-synchronized after warm-up and exclude model loading, video encoding, and
disk I/O.

\begin{table}[t]
\centering
\begingroup
\small
\setlength{\tabcolsep}{5.0pt}
\renewcommand{\arraystretch}{1.05}
\begin{tabular*}{\textwidth}{@{\extracolsep{\fill}}lccc@{}}
\toprule
Stage & Time (s) & Share & Peak memory \\
\midrule
Reference rollout & $72.6\pm7.5$ & 24.3\% & 30.5 GB \\
SAM~2 extraction  & $24.9\pm1.0$ & 8.3\%  & shared \\
Controlled rollout & $201.9\pm3.8$ & 67.4\% & 30.5 GB \\
\midrule
End-to-end & $\mathbf{299.4\pm10.0}$ & 100\% & 30.5 GB \\
Full Memory, one pass & $72.6\pm7.5$ & -- & 30.5 GB \\
\bottomrule
\end{tabular*}
\endgroup
\caption{\textbf{Runtime of the offline two-pass pipeline.} Mean $\pm$
standard deviation over seven approximately 30-second videos.}
\label{tab:offline_runtime}
\end{table}

The controlled rollout is $2.81\pm0.27\times$ the Full-Memory pass because the
current implementation constructs per-frame biases and evaluates separate
subject/background attention calls. The complete reference--segmentation--
controlled pipeline is $4.15\pm0.30\times$ a single Full-Memory rollout.
Online mask updates would require additional decoding and segmentation and are
not included in these measurements.

\subsection{Sensitivity to Subject-Prior Quality}
\label{sec:prior_sensitivity}

We test mask-boundary sensitivity on seven prompts with one seed. All
conditions share the checkpoint, initial noise, reference rollout, and routing
hyperparameters. We compare (i) the original latent-grid mask, (ii) a
coarse mask obtained by downsampling the $30\times52$ grid to $8\times13$ with
nearest-neighbor interpolation and upsampling it back, producing approximately
$64\times64$-pixel blocks, and (iii) the filled minimum enclosing box in every
frame. The coarse and box masks have mean IoU ranges of 0.57--0.76 and
0.67--0.84 against the original mask, respectively.

The coarse prior is close to the original in nT (1.463 vs.\ 1.490) and
tail-subject consistency (0.932 vs.\ 0.934). The box prior has higher nT but
lower tail-subject consistency and a higher artifact-veto rate. Table~\ref{tab:prior_sensitivity}
reports the complete comparison.

\begin{figure*}[!t]
  \centering
  \includegraphics[width=0.90\textwidth]{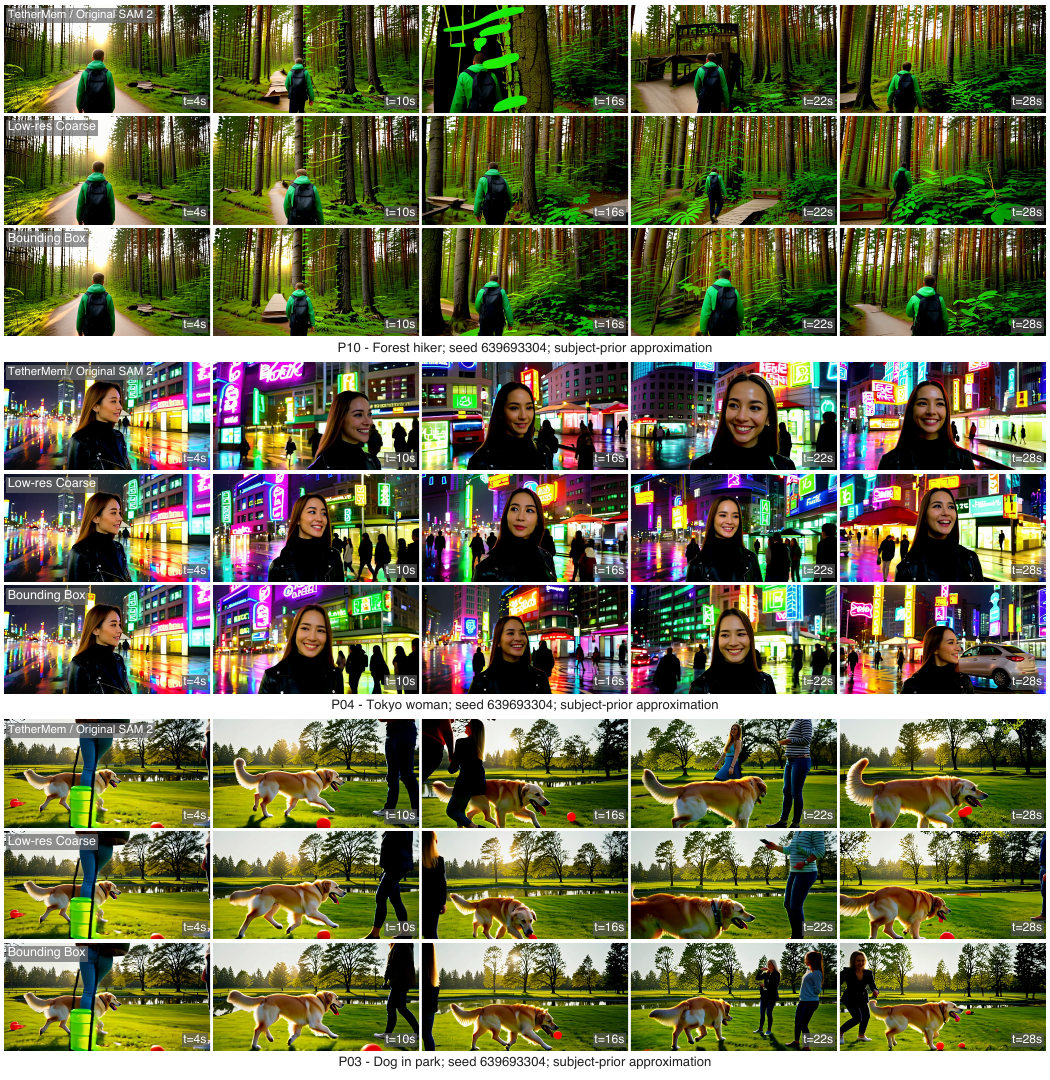}
  \caption{\textbf{Output sensitivity to subject-prior approximation.}
  Prompt blocks: P10, P04, and P03. Rows: original, coarse, and bounding-box
  priors. Columns: frames at 4, 10, 16, 22, and 28 seconds. All videos share
  the same seed and settings.}
\label{fig:supp_prior_output_robustness}
\end{figure*}

\section{Mechanism-Focused Ablations}

\subsection{Post-Attention Design Baseline}

We compare Full Memory and post-attention Value Reweighting directly with
TetherMem over seven prompts and two seeds. Three annotators judge each of the
28 pairs, giving 84 judgments. Ties are retained and ``unsure'' responses are
excluded.

All three variants use the same generator and inference settings. Full
Memory leaves retrieved-memory attention unchanged. Value Reweighting applies
the subject weight of one and the area-calibrated background weight $\gamma_n$
directly to historical Values after attention routing; it uses the current
query mask tiled over historical keys, with no query split or age prior.
TetherMem instead applies the complete normalized regional and recency-aware
routing in Table~\ref{tab:routing_configuration}.

Table~\ref{tab:value_reweight_ep}(b) reports regularized Davidson EP and
automatic metrics over the 14 prompt--seed cells. Value Reweighting approaches
TetherMem in nT but has lower human EP and tail-subject consistency.

\subsection{Full-Memory-Centered Routing Ablation}
\label{sec:full_centered_ablation}

The ablation compares three routing variants directly with Full Memory over
seven prompts and two seeds (14 matched units; three judgments per pair). Ties
are retained and ``unsure'' judgments are excluded. Routing w/o Region keeps
the age prior, Routing w/o Age keeps regional separation, and TetherMem combines
both factors.

\begin{table}[H]
\centering
\begingroup
\scriptsize
\textbf{(a) Direct outcomes against Full Memory}\par\smallskip
\begin{tabularx}{0.72\linewidth}{l*{4}{>{\centering\arraybackslash}X}}
\toprule
Variant & Ovr. & Prog. & ID & L1 \\
\midrule
Routing w/o Region & 14/16/12 & 15/17/10 & 12/18/11 & 14/17/11 \\
Routing w/o Age    & 15/15/12 & 12/20/10 & 12/19/10 & 14/16/12 \\
\rowcolor{claimblue}
TetherMem          & 25/6/11  & 29/5/8   & 18/9/14  & 19/10/13 \\
\bottomrule
\end{tabularx}

\medskip
\textbf{(b) Common-scale expected preference and nT}\par\smallskip
\begin{tabularx}{0.72\linewidth}{l*{5}{>{\centering\arraybackslash}X}}
\toprule
Variant & Ovr. & Prog. & ID & L1 & nT$\uparrow$ \\
\midrule
Full Memory        & 0.426 & 0.391 & 0.472 & 0.457 & 0.980 \\
Routing w/o Region & 0.461 & \underline{0.472} & 0.489 & \underline{0.505} & \underline{1.232} \\
Routing w/o Age    & \underline{0.476} & 0.428 & \underline{0.505} & 0.490 & 1.083 \\
\rowcolor{claimblue}
\textbf{TetherMem} & \textbf{0.638} & \textbf{0.709} & \textbf{0.534} & \textbf{0.548} & \textbf{1.458} \\
\bottomrule
\end{tabularx}
\endgroup
\caption{\textbf{Full-memory-centered routing ablation.}
\textbf{(a)} Direct win/tie/loss counts from each row variant's perspective.
\textbf{(b)} Regularized Davidson EP with nT averaged over two seeds.}
\label{tab:full_centered_wtl}
\label{tab:full_centered_ep}
\end{table}

TetherMem has the highest point estimate on all human dimensions and nT; its
intervals against the strongest single-factor variant include zero
(Table~\ref{tab:ablation}), so the factor ordering is directional.

\section{Limitations and Released Artifacts}
\label{sec:scope_traceability}

The evaluation uses ten prompts, three seeds, and one primary
Wan2.1-T2V-1.3B/LongLive-RAG stack. The Deep Forcing transfer remains within the
Wan2.1 family. TetherMem tracks at most one subject and requires an offline
reference rollout followed by SAM~2 mask extraction and controlled generation.
The reference masks become less aligned late in the rollout, and the human mask
comparison uses one annotator. Results beyond approximately 30 seconds are
qualitative examples.

\begin{figure*}[!p]
\centering
\includegraphics[width=\textwidth]{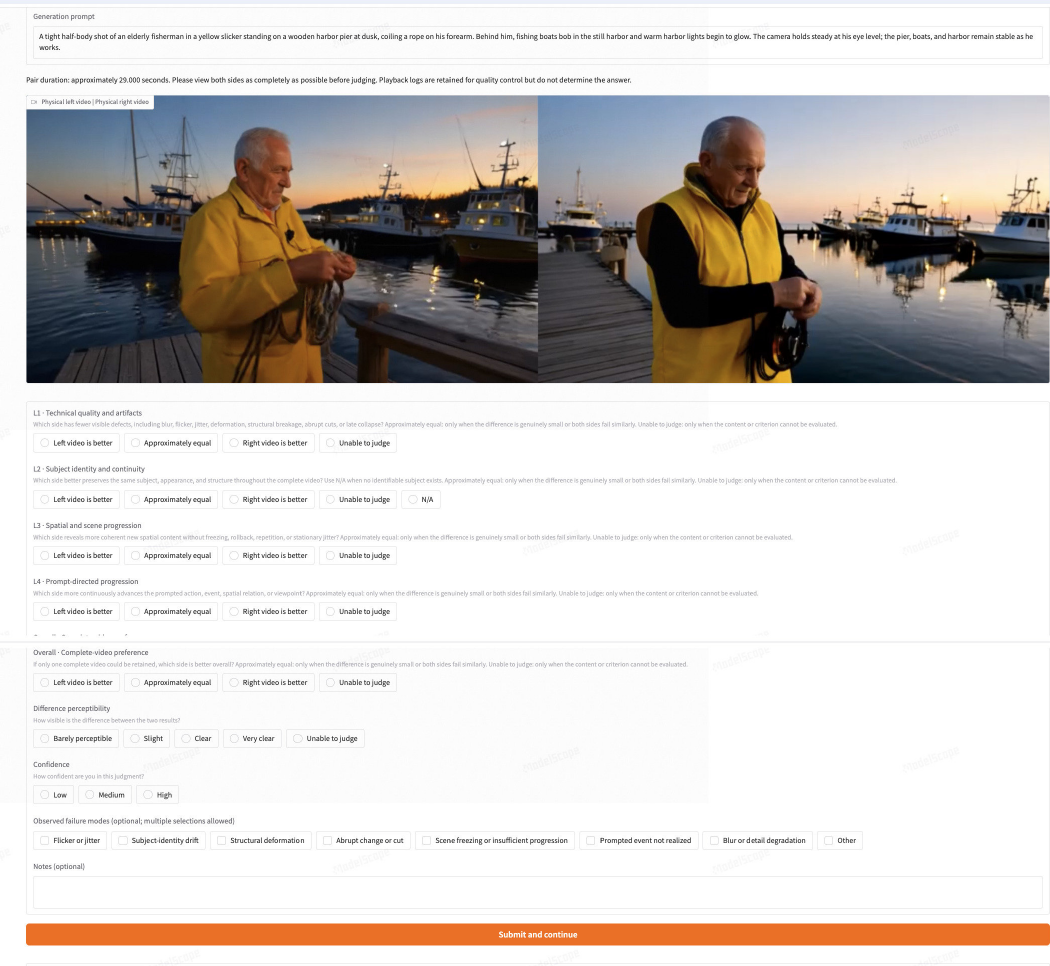}
\caption{\textbf{Participant-facing blind-evaluation form.}
The anonymized view shows the prompt, video pair, L1--L4 and Overall questions,
and optional diagnostics for one Table~\ref{tab:main_sota} task.}
\label{fig:annotation_interface_audit}
\end{figure*}

The Code and Data Supplement provides anonymized evaluation records, figure
data, and 15 compressed video demonstrations. Figure~\ref{fig:annotation_interface_audit}
shows the human-evaluation interface.

\FloatBarrier

\clearpage
\clearpage

\end{document}